%% file: Hyper_Spec_v6.tex
\documentclass[10pt,letterpaper, journal]{IEEEtran}
\pdfoutput=1
\usepackage{amsmath,graphicx} 
\usepackage{amsfonts, amssymb, amsthm, array}
\usepackage[T1]{fontenc}    
\usepackage{url}            
\usepackage{booktabs}       
\usepackage{nicefrac}       
\usepackage{microtype}      
\usepackage{bbm}
\usepackage{cite}
\input{commands}

\usepackage{epsfig}
\usepackage{multirow}
\newtheorem{theorem}{Theorem}[]
\newtheorem{lemma}[theorem]{Lemma}
\usepackage{color,soul}
\usepackage{subcaption}
\usepackage[font=small,labelfont=bf]{caption}
\usepackage{algorithm}
\usepackage{algorithmic, bm, textcomp}
\usepackage[normalem]{ulem}
\usepackage{tabularx, booktabs}
\usepackage{xargs}                      
\newcolumntype{P}[1]{>{\centering\arraybackslash}p{#1}}
\newcolumntype{Y}{>{\centering\arraybackslash}m{0.9cm}}
\newcolumntype{G}{>{\bfseries\centering\arraybackslash}m{1.8cm}}
\usepackage{etoolbox}{\tiny }
\usepackage[colorlinks,
linkcolor=blue,
anchorcolor=blue,
citecolor=blue
]{hyperref}

\usepackage[pdftex,dvipsnames, table]{xcolor} 
\usepackage[colorinlistoftodos]{todonotes}
\newcommandx{\unsure}[2][1=]{\todo[linecolor=red,backgroundcolor=red!25,bordercolor=red,#1]{#2}}
\newcommandx{\change}[2][1=]{\todo[linecolor=blue,backgroundcolor=blue!25,bordercolor=blue,#1]{#2}}
\newcommandx{\info}[2][1=]{\todo[linecolor=OliveGreen,backgroundcolor=OliveGreen!25,bordercolor=OliveGreen,#1]{#2}}
\newcommandx{\improvement}[2][1=]{\todo[linecolor=Plum,backgroundcolor=Plum!25,bordercolor=Plum,#1]{#2}}

\newcommand{\rk}{{\rm rank}}

\newcommand{\spann}{{\rm span}}

\newcommand{\csupp}{\text{csupp}}

\newcommand{\col}{\text{col}}
\newcommand{\RR}{\mathbb{R}}

\makeatletter
\@tempswafalse
\@ifpackageloaded{amsthm}{\@tempswatrue}{}
\@ifpackageloaded{ntheorem}{\@tempswatrue}{}
\if@tempswa
  \patchcmd\@thm{\trivlist}{\normalsize\trivlist}{}{}
\else
  \patchcmd\@begintheorem{\trivlist}{\normalsize\trivlist}{}{}
  \patchcmd\@opargbegintheorem{\trivlist}{\normalsize\trivlist}{}{}
\fi
\makeatother

\def\[#1\]{{$#1$}}
\def\@#1\@{{\mathbf{#1}}}
\def\b#1{{\mathbf{#1}}}
\def\c#1{{\mathcal{#1}}}

\makeatletter
\g@addto@macro \small {%
 \setlength\abovedisplayskip{5pt plus 2pt minus 2pt}%
 \setlength\belowdisplayskip{5pt plus 2pt minus 2pt}%
}
\makeatother

\makeatletter
\g@addto@macro \footnotesize {%
 \setlength\abovedisplayskip{5pt plus 3pt minus 3pt}%
 \setlength\belowdisplayskip{5pt plus 3pt minus 3pt}%
}
\makeatother

\makeatletter
\g@addto@macro \normalsize {%
 \setlength\abovedisplayskip{5pt plus 3pt minus 3pt}%
 \setlength\belowdisplayskip{5pt plus 3pt minus 3pt}%
}
\makeatother

\newcommand{\edit}[1]{\textcolor{black}{#1}}
\theoremstyle{definition}
\newtheorem{definition}{Definition D.\ignorespaces}[]

\newtheorem*{remark*}{Remark}

\IEEEoverridecommandlockouts

\usepackage{hyperref}
\title{A Dictionary-Based Generalization of Robust PCA Part II: Applications to Hyperspectral Demixing}
\author{\IEEEauthorblockN{Sirisha Rambhatla, Xingguo Li, Jineng Ren and Jarvis Haupt}\\
	\IEEEauthorblockA{Department of Electrical and Computer Engineering,\\ University of Minnesota -- Twin Cities, Minneapolis, MN-55455\\
		{\tt \{rambh002, lixx1661, renxx282, jdhaupt\}@umn.edu}. }}
	
	\author{Sirisha Rambhatla$^\dagger$,~\IEEEmembership{Student Member,~IEEE,}
		Xingguo Li$^\dagger$,~\IEEEmembership{Student Member,~IEEE,}
		Jineng Ren$^\dagger$,~\IEEEmembership{Student Member,~IEEE,}
		and~Jarvis Haupt$^\dagger$,~\IEEEmembership{Senior Member,~IEEE}\vspace{-15pt}
		 \thanks{This work was supported by the DARPA YFA, Grant N66001-14-1-4047. Preliminary versions appeared in the proceedings of the 2016 IEEE Global Conference on Signal  \& Information Processing (GlobalSIP), 2017 Asilomar Conference on Signals, Systems, \& Computers, and the 2018 IEEE International Conference on Acoustics, Speech \& Signal Processing (ICASSP).}
		\thanks{$^\dagger$ S. Rambhatla, J. Ren, and J. Haupt are with the Department
			of Electrical and Computer Engineering, University of Minnesota, Minneapolis,
			MN, 55455, USA e-mail: {\tt \{rambh002, renxx282, jdhaupt\}@umn.edu}, respectively. X. Li is with the Computer Science Department, Princeton University, Princeton, NJ 08540, USA email: {\tt xingguol@cs.princeton.edu}. Different authors contributed to different phases of the work.}
		 }

\begin{document}
%
\maketitle
\begin{abstract}
We consider the task of localizing targets of interest in a hyperspectral (HS) image based on their spectral signature(s), by posing the problem as two distinct convex demixing task(s).
With applications ranging from remote sensing to surveillance, this task of target detection leverages the fact that each material/object possesses its own characteristic spectral response, depending upon its composition. 
However, since \textit{signatures} of different materials are often correlated, matched filtering-based approaches may not be apply here.  To this end, we model a HS image as a superposition of a low-rank component and a dictionary sparse component, wherein the dictionary consists of the \textit{a priori} known characteristic spectral responses of the target we wish to localize, and develop techniques for two different sparsity structures, resulting from different model assumptions. We also present the corresponding recovery guarantees, leveraging our recent theoretical results from a companion paper. Finally, we analyze the performance of the proposed approach via experimental evaluations on real HS datasets for a classification task, and compare its performance with related techniques.


\end{abstract}
\vspace*{-5pt}
\section{Introduction}
\label{sec:intro}
Hyperspectral (HS) imaging is an imaging modality which senses the intensities of the reflected electromagnetic waves (responses) corresponding to different wavelengths of the electromagnetic spectra, often invisible to the human eye. As the spectral response associated with an object/material is dependent on its composition, HS imaging lends itself very useful in identifying the said target objects/materials via their characteristic spectra or  \textit{signature} responses, also referred to as \emph{endmembers} in the literature. 
Typical applications of HS imaging range from monitoring agricultural use of land, catchment areas of rivers and water bodies, food processing and surveillance, to detecting various minerals, chemicals, and even presence of life sustaining compounds on distant planets; see \cite{Borengasser2007, Park2015}, and references therein for details. However, often, these spectral \textit{signatures} are highly correlated, making it difficult to detect regions of interest based on these endmembers. In this work, we present two techniques to localize target materials/objects in a given HS image based on some structural assumptions on the data, using the \textit{a priori} known signatures of the target of interest.

The primary property that enables us to localize a target is the approximate low-rankness of HS images when represented as a matrix, owing to the fact that a particular scene is composed of only a limited type of objects/materials \cite{Keshava2002}. For instance, while imaging an agricultural area, one would expect to record responses from materials like biomass, farm vehicles, roads, houses, water bodies, and so on. Moreover, the spectra of complex materials can be assumed to be a linear mixture of the constituent materials \cite{Keshava2002, Greer2012}, i.e. the received HS responses can be viewed as being generated by a linear mixture model \cite{Xing2012}. 
For the target localization task at hand, this approximate low-rank structure is used to decompose a given HS image into a low-rank part, and a component that is sparse in a known dictionary -- a \textit{dictionary sparse} part-- wherein the dictionary is composed of the spectral signatures of the target of interest. We begin by formalizing the specific model of interest in the next section.  

\vspace*{-7pt}
\subsection{Model}
\begin{figure}[t]
  \centering
  \begin{tabular}{c}
    \includegraphics[width=0.25\textwidth]{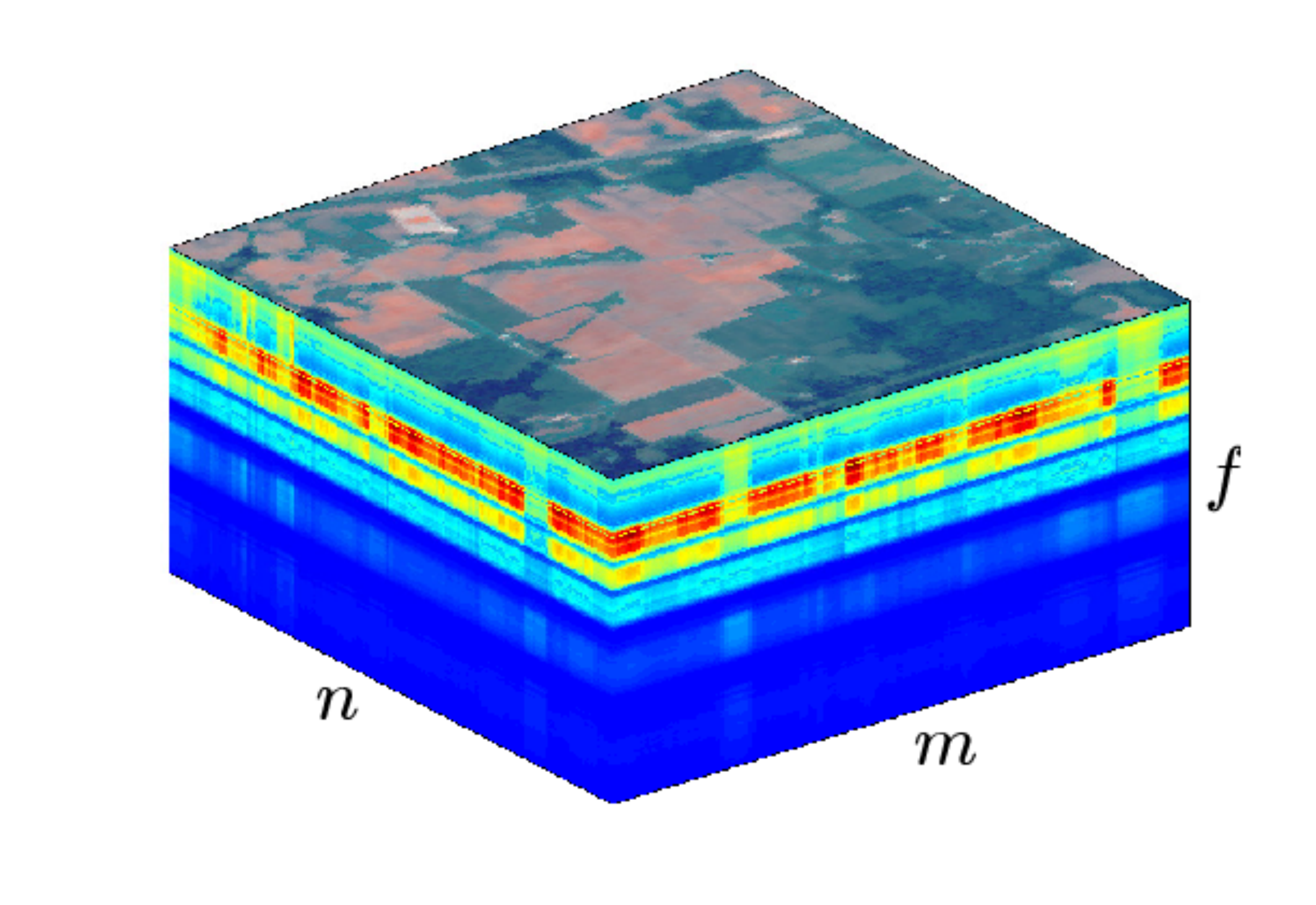}
    \end{tabular}
  \caption{ The HS image data-cube corresponding to the Indian Pines dataset.} 
  \label{fig:data} 
  \vspace{-15pt}
\end{figure}

A HS sensor records the response of a region, which corresponds to a pixel in the HS image as shown in Fig.~\ref{fig:data}, to different frequencies of the electromagnetic spectrum. As a result, each HS image $\mb{I} \in \mathbb{R}^{n \times m \times f}$, can be viewed as a data-cube formed by stacking $f$ matrices of size {$n \times m$}, as shown in Fig.~\ref{fig:data}. Therefore, each volumetric element or \textit{voxel}, of a HS image is a vector of length $f$, and represents the response of the material to $f$ measurement channels. Here,  $f$ is determined by the number of channels or frequency bands across which measurements of the reflectances are made.

%

Formally, let {$\mb{M}\in\mathbb{R}^{ f \times nm}$} be formed by \textit{unfolding} the HS image {$\mb{I}$}, such that, each column of {$\mb{M}$} corresponds to a voxel of the data-cube. We then model {$\mb{M}$} as arising from a superposition of a low-rank component {$\mb{L} \in \mathbb{R}^{f \times nm}$} with rank { $r$}, and a dictionary-sparse component, expressed as {$\mb{DS}$}, i.e.,
	\begin{align}
	\label{Prob}
	\mb{M} = \mb{L} + \mb{DS}.
	\end{align}%
Here, {$\mb{D} \in \mathbb{R}^{f \times d}$} represents an \textit{a priori} known dictionary composed of appropriately normalized characteristic responses of the material/object (or the constituents of the material), we wish to \edit{localize}, and {$\mb{S} \in \mathbb{R}^{d \times nm}$} refers to the \textit{sparse} coefficient matrix (also referred to as \emph{abundances} in the literature). 
Note that {$\mb{D}$} can also be constructed by learning a dictionary based on the known spectral signatures of a target; see \cite{Olshausen97,Aharon05, Mairal10,Lee2007}. 
\vspace*{-5pt}
\subsection{Our Contributions}\label{sec:sp_choice}
In this work, we present two techniques\footnote{The code is made available at \texttt{github.com/srambhatla/Diction ary-based-Robust-PCA}.} for target detection in a HS image, depending upon different sparsity assumptions on the matrix \[\b{S}\], by modeling the data as shown in \eqref{Prob}. 
Building on the theoretical results of \cite{Rambhatla2016, Li2018, Rambhatla18LrTheo}, our techniques operate by forming the dictionary \[\b{D}\] using the \textit{a priori} known spectral signatures of the target of interest, and leveraging the approximate low-rank structure of the data matrix \[\b{M}\] \cite{Rambhatla17}. Here, the dictionary \[\b{D}\] can be formed from the \textit{a priori} known signatures directly, or by learning an appropriate dictionary based on target data; see \cite{Olshausen97,Aharon05, Mairal10,Lee2007}. 
\begin{figure}[t]
  \centering
  \begin{tabular}{c}
    \includegraphics[width=0.4\textwidth]{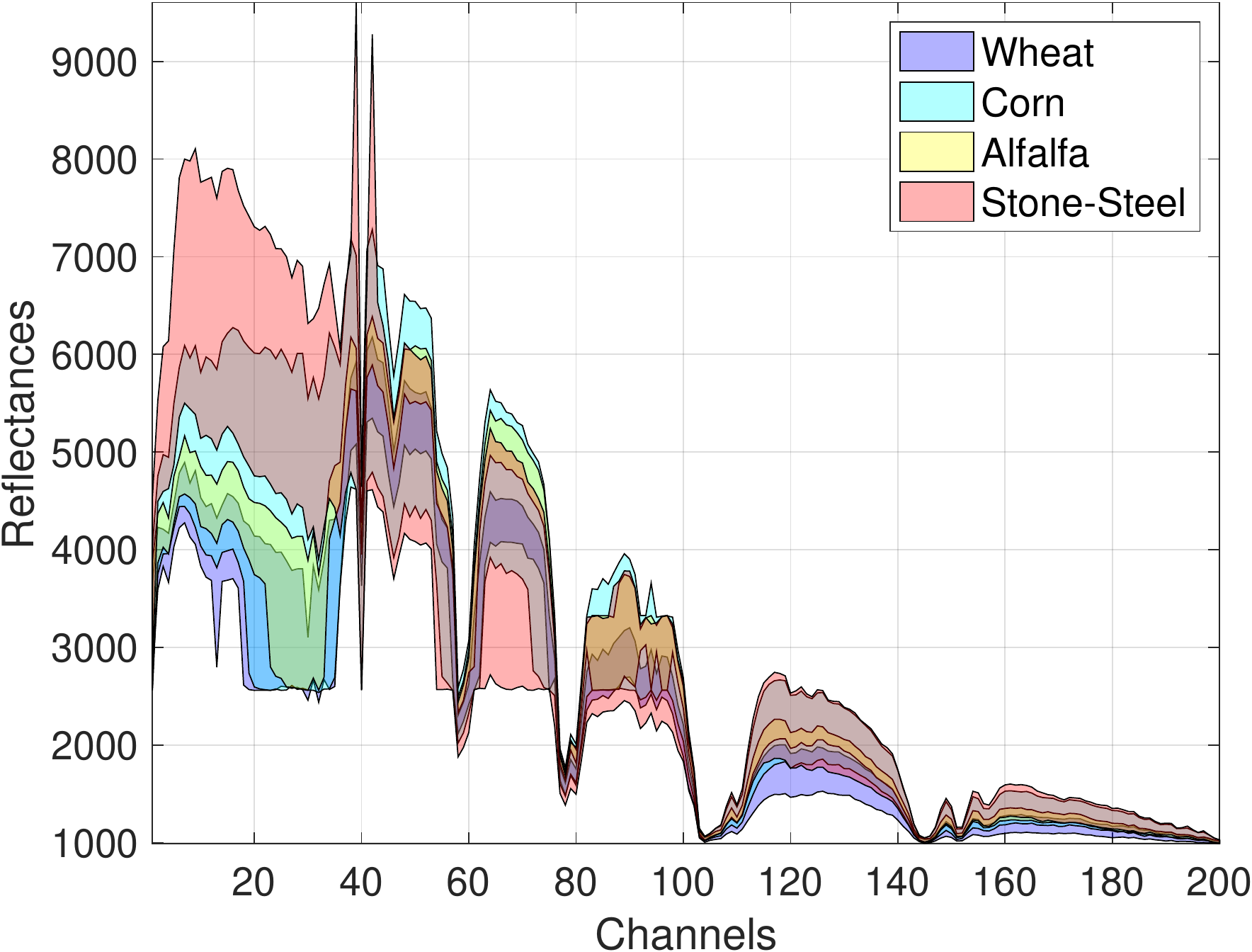}
    \end{tabular}
  \caption{Correlated spectral signatures. The spectral signatures of even different materials are highly correlated. Shown here are spectral signatures of classes from the Indian Pines dataset \cite{HSdat}. Here, the shaded region shows the lower and upper ranges of reflectance values the signatures take. }
  \label{fig:data_corr} 
  \vspace{-12pt}
\end{figure}

We consider two types of sparsity structures for the coefficient matrix \[\b{S}\], namely, a) \textit{global} or \textit{entry-wise} sparsity, wherein we let the matrix \[\b{S}\] have \[s_e\] non-zero entries globally, and b) \textit{column-wise} sparse structure, where at most \[s_c\] columns of the matrix \[\b{S}\] have non-zero elements. The choice of a particular sparsity model depends on the properties of the dictionary matrix \[\b{D}\]. In particular, if the target signature admits a sparse representation in the dictionary, entry-wise sparsity structure is preferred. This is likely to be the case when the dictionary is overcomplete (\[f<d\]) or \textit{fat}, and also when the target spectral responses admit a sparse representation in the dictionary. On the other hand, the column-wise sparsity structure is amenable to cases where the representation can use all columns of the dictionary. This potentially arises in the cases when the dictionary is undercomplete (\[f\geq d\]) or \textit{thin}. Note that, in the column-wise sparsity case, the non-zero columns need not be sparse themselves. The applicability of these two modalities is also exhibited in our experimental analysis; see Section~\ref{sec:exp} for further details. Further, we specialize the theoretical results of  \cite{Rambhatla18LrTheo}, to present the conditions under which such a demixing task will succeed under the two sparsity models discussed above; see also \cite{Rambhatla2016} and \cite{Li2018}.


Next, we analyze the performance of the proposed techniques via extensive experimental evaluations on real-world demixing tasks over different datasets and dictionary choices, and compare the performance of the proposed techniques with related works. This demixing task is particularly challenging since the spectral signatures of even distinct classes are highly correlated to each other, as shown in Fig.~\ref{fig:data_corr}. The shaded region here shows the upper and lower ranges of different classes. For instance, in Fig.~\ref{fig:data_corr} we observe that the spectral signature of the ``Stone-Steel'' class is similar to that of class ``Wheat''. This correlation between the spectral signatures of different classes results in an approximate low-rank structure of the data, captured by the low-rank component \[\b{L}\], while the dictionary-sparse component \[\b{DS}\] is used to identify the target of interest. We specifically show that such a  decomposition successfully localizes the target despite the high correlation between spectral signatures of distinct classes.

Finally, it is worth noting that although we consider \textit{thin} dictionaries (\[f\geq d\]) for the purposes of this work, since it is more suitable for the current exposition, our theoretical results are also applicable for the \textit{fat} case (\[f< d\]); see \cite{Rambhatla2016},\cite{Li2018}, and \cite{Rambhatla18LrTheo} for further details.

\vspace*{-5pt}
\subsection{Prior Art}
\label{priorart}
The model shown in \eqref{Prob} is closely related to a number of well-known problems. To start, in the absence of the dictionary sparse part  \[\b{D}\b{S}\], \eqref{Prob} reduces to the popular problem of principal component analysis (PCA) \cite{Pearson1901,Jolliffe02}. The problem considered here also shares its structure with variants of PCA, such as robust-PCA \cite{Candes11, Chandrasekaran11} (with \[\b{D} = \b{I}\] for an identity matrix \[\b{I}\],) outlier pursuit \cite{Xu2010} (where \[\b{D} = \b{I}\] and \[\b{S}\] is column-wise sparse,) and others  \cite{Zhou2010, Ding11, Wright13, Chen13, Li15, Li15c, Li15b, Li16_refined, Li2016efficient}. 

On the other hand, the problem can be identified as that of sparse recovery \cite{Natarajan95, Donoho01, Candes05, Rambhatla2013}, in the absence of the low-rank part \[\b{L}\]. Following which, sparse recovery methods for analysis of HS images have been explored in \cite{Moudden2009, Bobin2009, Kawakami2011, Charles2011}. In addition, in a recent work \cite{Giampouras2016}, the authors further impose a low-rank constraint on the coefficient matrix \[\b{S}\] for the demixing task. Further, applications of compressive sampling have been explored in \cite{Golbabaee2010}, while \cite{Xing2012} analyzes the case where HS images are noisy and incomplete. The techniques discussed above focus on identifying all materials in a given HS image. However, for target localization tasks, it is of interest to identify only specific target(s) in a given HS image. As a result, there is a need for techniques which localize targets based on their \textit{a priori} known spectral signatures.

The model described in \eqref{Prob} was introduced in \cite{Mardani2012} as a means to detect traffic anomalies in a network, wherein, the authors focus on a case where the dictionary \[\b{D}\] is \textit{overcomplete}, i.e., \textit{fat}, and the rows of \[\b{D}\] are orthogonal, e.g., $\mb{RR^\top} = \mb{I}$. Here, the coefficient matrix \[\b{S}\] is assumed to possess at most $k$ nonzero elements per row and column, and $s$ nonzero elements globally. In a recent work \cite{Rambhatla2016} and the accompanying theoretical work \cite{Rambhatla18LrTheo}, we analyze the extension of \cite{Mardani2012} to include a case where the dictionary has more rows than columns, i.e., is \textit{thin}, while removing the orthogonality constraint for both the \textit{thin} and the \textit{fat} dictionary cases, when {$s$} is small. This case is particularly amenable for the target localization task at hand, since often we aim to localize targets based on a few \textit{a priori} known spectral signatures. 
To this end, we focus our attention on the \textit{thin} case, although a similar analysis applies for the \textit{fat} case \cite{Rambhatla2016}; see also \cite{Rambhatla18LrTheo}.

\vspace*{-5pt}
\subsection{Related Techniques}
\label{rel_tech}
To study the properties of our techniques, we compare and contrast their performance with related works. First, as a sanity check, we compare the performance of the proposed techniques with matched filtering-based methods (detailed in Section~\ref{sec:exp}). In addition, we compare the performance of our techniques to other closely related methods based on the sparsity assumptions on the matrix \[\b{S}\], as described below.

\vspace{6pt}
\noindent\textbf{For entry-wise sparse structure:} The first method we compare to is based on the observation that in cases where the known dictionary $\mb{D}$ is thin, we can multiply \eqref{Prob} on the left by the pseudo-inverse of $\mb{D}$, say $\b{D}^\dagger$, in which case, the model shown in \eqref{Prob} reduces to that of robust PCA, i.e.,
	\begin{align}
	\tag{RPCA$^\dagger$}
	\label{Prob_rpca}
	\widetilde{\b{M}} = \widetilde{\b{L}} + \b{S},
	\end{align}%
where $\mb{\widetilde{M}} = \mb{D}^\dagger\mb{M}$ and $\widetilde{\b{L}} = \mb{D}^\dagger\mb{L}$. Therefore, in this case, we can recover the sparse matrix $\mb{S}$ by robust PCA \cite{Candes11, Chandrasekaran11}, and estimate the low-rank part using the estimate of $\mb{DS}$. Note that this is not applicable for the \textit{fat} case due to the non-trivial null space of its pseudo-inverse.

Although at a first glance this seems like a reasonable technique, somewhat surprisingly, it does not succeed for all \textit{thin} dictionaries. Specifically, in cases where \[r\], the rank of \[\b{L}\], is greater than the number of dictionary elements \[d\], the pseudo-inversed component \[\widetilde{\b{L}}\] is no longer ``low-rank.'' In fact, since the notion of low-rankness is relative to the potential maximum rank of the component, \[\widetilde{\b{L}}\] can be close to full-rank. As a result, the robust PCA model shown in \ref{Prob_rpca} is no longer applicable and the demixing task may not succeed; see our corresponding theoretical paper \cite{Rambhatla18LrTheo} for details. 

Moreover, even in cases where \ref{Prob_rpca} succeeds (\[r < d\]), our proposed one-shot procedure guarantees the recovery of the two components under some mild conditions, while the pseudo-inverse based procedure \ref{Prob_rpca} will require a two-step procedure -- one to recover the sparse coefficient matrix and other to recover the low-rank component -- in addition to a non-trivial analysis of the interaction between $\mb{D}^\dagger$ and the low-rank part \[\b{L}\]. This is also apparent from our experiments shown in Section~\ref{sec:exp}, which indicate that optimization based on the model in \eqref{Prob} is more \textit{robust} as compared to \ref{Prob_rpca} for the classification problem at hand across different choices of the dictionaries.

\vspace{6pt}
\noindent\textbf{For column-wise sparse structure:} The column-wise sparse structure of the matrix \[\b{S}\] results in a column-wise sparse structure of the dictionary-sparse component \[\b{DS}\]. As a result, the model at hand is similar to that studied in OP \cite{Xu2010}. Specifically, the OP technique is aimed at identifying the outlier columns in a given matrix. However, it fails in cases where the target of interest is not an outlier, as in case of HS data. On the other hand, since the proposed technique uses the dictionary \[\b{D}\] corresponding to the spectral signatures of the target of interest to guide the demixing procedure, it results in a spectral signature-driven technique for target localization. This distinction between the two procedures is also discussed in our corresponding theoretical work \cite{Rambhatla18LrTheo} Section V, and is exemplified by our experimental results shown in Section~\ref{sec:exp}. 



Further, as in the entry-wise case, one can also envision a pseudo-inverse based procedure to identify the target of interest via OP \cite{Xu2010} on the pseudo-inversed data (referred to as \ref{Prob_op} in our discussion) i.e., 
	\begin{align}
	\tag{OP$^\dagger$}
	\label{Prob_op}
	\widetilde{\b{M}} = \widetilde{\b{L}} + \mb{S},
	\end{align}%
where $\widetilde{\b{M}} = \mb{D}^\dagger\mb{M}$ and $\widetilde{\b{L}} = \mb{D}^\dagger\mb{L}$, with \[\b{S}\] admitting a column-wise sparse structure. However, this variant of OP does not succeed when the rank of the low-rank component is greater than the number of dictionary elements, i.e., \[r\geq d\], as in the previous case; see our related theoretical work for details \cite{Rambhatla18LrTheo} Section V.


The rest of the paper is organized as follows. We formulate the problem and introduce relevant theoretical quantities in Section~\ref{sec:prelim}, followed by specializing the theoretical results for the current application in Section~\ref{sec:main_result}. Next, in Section~\ref{sec:optimization}, we present the specifics of the algorithms for the two cases. In Section~\ref{sec:exp}, we describe the experimental set-up and demonstrate the applicability of the proposed approaches via extensive numerical simulations on real HS datasets for a classification task. Finally, we conclude this discussion in Section~\ref{sec:conclusion}.

\vspace{6pt}
\noindent\textbf{Notation:} Given a matrix \[\b{X}\], \[\b{X}_i\] denotes its $i$-th column and \[\b{X}_{i,j}\] denotes the \[(i,j)\] element of \[\b{X}\]. We use \[\|\b{X}\|:= \sigma_{\max}(\b{X})\] for the spectral norm, where \[\sigma_{\max}(\b{X})\] denotes the maximum singular value of the matrix, \[\|\b{X}\|_\infty := \underset{i,~j}{\max} |\b{X}_{ij}|\], \[\|\b{X}\|_{\infty, \infty} :=
\underset{i}{\max} \|\b{e}^\top_i\b{X}\|_1 \], and \[\|\b{X}\|_{\infty,2} := \underset{i}{\max} \|\b{X}\b{e}_i\|\], where \[\b{e}_i\] denotes the canonical basis vector with \[1\] at the \[i\]-th location and \[0\] elsewhere. Further, we denote the \[\ell_2\]-norm of a vector \[\b{x}\] as \[\|\b{x}\|\].  In addition, \[\|.\|_* \], \[\|.\|_1\], and \[\|.\|_{1,2}\] refer to the nuclear norm, entry-wise \[\ell_1\]- norm, and \[\ell_{1,2}\] norm (sum of the \[\ell_2\] norms of the columns) of the matrix, respectively, which serve as convex relaxations of rank, sparsity, and column-wise sparsity inducing optimization, respectively.

\section{Problem Formulation} \label{sec:prelim}
In this section, we introduce the optimization problem of interest and different theoretical quantities pertinent to our analysis. 

\vspace*{-5pt}
\subsection{Optimization problems}

Our aim is to recover the low-rank component  \[\b{L}\] and the sparse coefficient matrix \[\b{S}\], given the dictionary \[\b{D}\] and samples {$\mb{M}$} generated according to the model shown in \eqref{Prob}. Here the coefficient matrix \[\b{S}\] can either have an entry-wise sparse structure or a column-wise sparse structure. 
We now crystallize our model assumptions to formulate appropriate convex optimization problems for the two sparsity structures. 

Specifically, depending upon the priors about the sparsity structure of  \[\b{S}\], and the low-rank property of the component \[\b{L}\], we aim to solve the following convex optimization problems, i.e.,
\begin{align}\tag{{D-RPCA(E)}}
\underset{\b{L}, \b{S}}{\min~} \|\b{L}\|_* + \lambda_e \|\b{S}\|_1 ~~\text{s.t.}~~ \b{M} = \b{L} + \b{DS} \label{Pe}
\end{align}
for the entry-wise sparsity case, and
\begin{align}\tag{{D-RPCA(C)}}
\underset{\b{L}, \b{S}}{\min~} \|\b{L}\|_* + \lambda_c \|\b{S}\|_{1,2} ~\text{s.t.}~\b{M} = \b{L} + \b{DS} \label{Pc}
\end{align}
for the column-wise sparse case, to recover \[\b{L}\] and \[\b{S}\] with regularization parameters \[\lambda_e\geq 0\] and \[\lambda_c \geq 0\], given the data \[\b{M}\] and the dictionary \[\b{D}\]. Here, the \textit{a priori} known dictionary  \[\b{D}\] is assumed to be  undercomplete (\textit{thin}, i.e., \[d\leq f\]) for the application at hand. Analysis of a more general case can be found in\cite{Rambhatla18LrTheo}. Further, here ``D-RPCA'' refers to ``dictionary based robust principal component analysis'', while the qualifiers ``E'' and ``C'' indicate the entry-wise and column-wise sparsity patterns, respectively.

Note that, in the column-wise sparse case there is an inherent ambiguity regarding the recovery of the true component pairs \[(\b{L}, \b{S})\] corresponding to the low-rank part and the dictionary sparse component, respectively.  Specifically, any pair \[(\b{L}_0, \b{S}_0)\] satisfying \[\b{M} = \b{L}_0 + \b{D}\b{S}_0 = \b{L}+ \b{D}\b{S}\], where \[\b{L}_0\] and \[\b{L}\] have the same column space, and \[\b{S}_0\] and \[\b{S}\] have the identical column support, is a solution of \ref{Pc}. To this end, we define the following \textit{oracle model} to characterize the optimality of any solution pair \[(\b{L}_0, \b{S}_0)\]. 
\begin{definition}[Oracle Model for Column-wise Sparse Case]\label{def:oracle}
\vspace{-2pt}
	Let the pair \[(\b{L}, \b{S})\] be the matrices forming the data \[\b{M}\] as per \eqref{Prob}, define the corresponding oracle model \[\{ \b{M}, \c{U}, \c{I}_{\c{S}_c}  \}\]. Then, any pair \[(\b{L}_0, \b{S}_0)\] is in the \emph{Oracle Model} \[\{ \b{M}, \c{U}, \c{I}_{\c{S}_c}  \}\], if \[\c{P}_{\c{U}}(\b{L}_0) = \b{L}\], \[\c{P}_{\c{S}_c}(\b{D} \b{S}_0) = \b{D} \b{S}\] and \[\b{L}_0 + \b{D} \b{S}_0 = \b{L} + \b{D} \b{S} = \b{M}\] hold simultaneously, where \[\c{P}_{\c{U}}\] and \[\c{P}_{\c{S}_c}\] are projections onto the column space \[\c{U}\] of \[\b{L}\] and column support \[\c{I}_{\c{S}_c} \] of \[\b{S}\], respectively.
	\vspace{-2pt}
\end{definition}
For this case, we then first establish the sufficient conditions for the existence of a solution based on some incoherence conditions. Following which, our main result for the column-wise case states the sufficient conditions under which solving a convex optimization problem recovers a solution pair \[(\b{L}_0, \b{S}_0)\] in the oracle model.


 \vspace*{-5pt}
\subsection{Conditions on the Dictionary}
For our analysis, we require that the dictionary \[\b{D}\] follows the \textit{generalized frame property} (GFP) defined as follows. 
\begin{definition}\label{frame}
\vspace{-2pt}
	A matrix \[\b{D}\] satisfies the \emph{generalized frame property} (GFP), on vectors \[\b{v} \in \c{R}\], if for any fixed vector \[\b{v}\in \c{R}\] where \[\b{v}\neq \b{0}\], we have
	\begin{align*}
	\alpha_\ell\|\b{v}\|^2_2 \leq \|\b{Dv}\|^2_2 \leq \alpha_u\|\b{v}\|^2_2, 
	\end{align*}
	where \[\alpha_\ell\] and \[\alpha_u\] are the lower and upper \emph{generalized frame bounds} with \[0 < \alpha_\ell \leq \alpha_u < \infty \].
	\vspace{-2pt}
\end{definition}
{The GFP is met as long as the vector \[\b{v}\] is not in the null-space of the matrix \[\b{D}\], and \[ \|\b{D}\| \] is bounded.  Therefore, for the \textit{thin} dictionary setting \[d < n\] for both entry-wise and column-wise sparsity cases, this condition is satisfied as long as \[\b{D}\] has a full column rank, and \[\c{R}\] can be the entire space. For example,  \[\b{D}\] being a \textit{frame}\cite{Duffin1952} suffices; see \cite{Heil2013} for a brief overview of frames. 
\vspace*{-5pt}
\subsection{Relevant Subspaces}

Before we define the relevant subspaces for this discussion, we define a few preliminaries. First, let the pair \[(\b{L_0}, \b{S_0})\] be the solution to \ref{Pe} (the entry-wise sparse case), and for the column-wise sparse case, let the pair \[(\b{L_0}, \b{S_0})\] be in the oracle model \[\{ \b{M}, \c{U}, \c{I}_{\c{S}_c} \}\]; see Definition \textcolor{blue}{\textbf{D.\ref{def:oracle}}}.
 
 Next, for the low-rank matrix \[\b{L}\], let the compact singular value decomposition (SVD) be represented as
	\begin{align*}
	\b{L} = \b{U\Sigma V^\top},
	\end{align*}
where \[\b{U}\in\RR^{f \times r}\] and \[\b{V}\in\RR^{nm \times r}\] are the left and right singular vectors of \[\b{L}\], respectively, and \[\b{\Sigma}\] is a diagonal matrix with singular values arranged in a descending order on the diagonal. Here, matrices \[\b{U}\] and \[\b{V}\] each have orthogonal columns. Further, let \[\c{L}\] be the linear subspace consisting of matrices spanning the same row or column space as \[\b{L}\], i.e.,
\begin{align*}
\c{L} := \{ \b{UW}_1^\top + \b{W}_2\b{V}^\top, \b{W}_1  \in \RR^{nm \times r}, \b{W}_2 \in \RR^{f \times r}\}.
\end{align*}

Next, let \[\c{S}_e\] (\[\c{S}_c\]) be the space spanned by \[d \times nm\] matrices with the same non-zero support (column support, denoted as $\csupp$) as \[\b{S}\], and let \[\c{D}\] be defined as
\begin{align*}
\c{D} := \{ \b{DH}\},\text{where}\begin{cases}
\b{H} \in \c{S}_e~\text{for entry-wise case},\\
\csupp(\b{H}) \subseteq \c{I}_{\c{S}_c}~\text{for column-wise case}.
\end{cases}
\end{align*}
Here, \[\c{I}_{\c{S}_c}\] denotes the index set containing the non-zero column indices of \[\b{S}\] for the column-wise sparsity case. In addition, we denote the corresponding complements of the spaces described above by appending `$\perp$'.  

We use calligraphic `\[\c{P}(\cdot)\]' to denote the projection operator onto a subspace defined by the subscript, and `\[\b{P}\]' to denote the corresponding projection matrix with the appropriate subscripts. Therefore, using these definitions the projection operators onto and orthogonal to the subspace \[\c{L}\] are defined as 
\begin{align*}
\c{P}_{\c{L}} (\b{L}) &= \b{P_U}\b{L} + \b{L} \b{P_V} -   \b{P_U}\b{L}\b{P_V}
\end{align*}
and
\vspace*{-3pt}
\begin{align*}
\c{P}_{\c{L}^\perp}(\b{L}) &=  (\b{I} - \b{P_U})\b{L}  (\b{I} - \b{P_V}),
\end{align*}
 respectively.
\vspace*{-8pt}
\subsection{Incoherence Measures}

We also employ various notions of incoherence to identify the conditions under which our procedures succeed. To this end, we first define the incoherence parameter \[\mu\] that characterizes the relationship between the low-rank part \[\b{L}\] and the dictionary sparse part \[\b{DS}\], as
\begin{align} \label{eqn:mu}
\mu := \underset{\b{Z} \in \mathcal{D} \backslash \{\b{0}_{d \times nm}\}}{\max} \tfrac{\|\mathcal{P}_{\mathcal{L}}(\b{Z})\|_{\rm F}}{\|\b{Z}\|_{\rm F}}.
\end{align}

The parameter \[\mu \in [0, 1]\] is the measure of degree of similarity between the low-rank part and the dictionary sparse component. Here, a larger \[\mu\] implies that the dictionary sparse component is close to the low-rank part.
In addition, we also define the parameter \[\beta_U\] as
\begin{align}\label{eqn:beta}
\b{\beta}_{\b{U}}  :=  \underset{\|\b{u}\| = 1}{\max} \tfrac{\|(\b{I} - \b{P}_{\b{U}}) \b{D}\b{u}\|^2}{\|\b{Du}\|^2},
\end{align}
which measures the similarity between the orthogonal complement of the column-space \[\c{U}\] and the dictionary \[\b{D}\].

The next two measures of incoherence can be interpreted as a way to identify the cases where for \[\b{L}\] with SVD as \[ \b{L} = \b{U \Sigma V^\top}\]: (a) \[\b{U}\] resembles the dictionary  \[\b{D}\], and (b) \[\b{V}\] resembles the sparse coefficient matrix \[\b{S}\]. In these cases, the low-rank part may resemble the dictionary sparse component. To this end, similar to \cite{Mardani2012}, we define the following measures to identify these cases as
\begin{align}\label{eqn:gamma}
\text{(a)} ~\gamma_{\b{U}}  :=  \underset{i}{\max} \tfrac{\|\b{P}_\b{U} \b{D}\b{e}_{i}\|^2}{\|\b{De}_{i}\|^2} ~{\normalsize\text{and}}~
\text{(b)}~ \gamma_{\b{V}} := \underset{i}{\max} \|\mathbf{P}_{\b{V}}\mathbf{e}_{i}\|^2.
\end{align}
Here, \[0 \leq \gamma_{\b{U}} \leq 1\] achieves the upper bound when a dictionary element is exactly aligned with the column space \[\c{U}\] of the \[\b{L}\], and lower bound when all of the dictionary elements are orthogonal to \[\c{U}\]. Moreover, \[\b{\gamma}_{\b{V}}  {\in [r/nm, 1]}\] achieves the upper bound when the row-space of \[\b{L}\] is ``spiky'', i.e., a certain row of \[\b{V}\] is \[1\]-sparse, meaning that a column of \[\b{L}\] is supported by (can be expressed as a linear combination of) a column of \[\b{U}\]. The lower bound here is attained when it is ``spread-out'', i.e., each column of \[\b{L}\] is a linear combination of all columns of \[\b{U}\]. In general, our recovery of the two components is easier when the incoherence parameters \[\gamma_{\b{U}}\] and \[\b{\gamma}_{\b{V}}\] are closer to their lower bounds.
In addition, for notational convenience, we define constants
\begin{align}\label{eqn:xi}
\xi_e := \|\b{D}^\top \b{UV}^\top\|_\infty ~~\text{and}~
\xi_c &:= \|\b{D}^\top \b{UV}^\top\|_{\infty,2}.
\end{align}
Here, \[\xi_e\] is the maximum absolute entry of \[\b{D}^\top \b{UV}^\top\], which measures how close columns of \[\b{D}\] are to the singular vectors of $\b{L}$. Similarly, for the column-wise case, \[\xi_c\] measures the closeness of columns of \[\b{D}\] to the singular vectors of $\b{L}$ under a different metric (column-wise maximal \[\ell_2\]-norm). 


\section{Theoretical Results }
\label{sec:main_result}
In this section, we specialize our theoretical results \cite{Rambhatla18LrTheo} for the HS demixing task. Specifically, we provide the main results corresponding to each sparsity structure of $\b{S}$ for the thin dictionary case considered here. We start with the theoretical results for the entry-wise sparsity case, and then present the corresponding theoretical guarantees for the column-wise sparsity structure; see \cite{Rambhatla18LrTheo} for detailed proofs.
\vspace*{-5pt}
\subsection{Exact Recovery for Entry-wise Sparsity Case}
For the entry-wise case, our main result establishes the existence of a regularization parameter \[\lambda_e\], for which solving the optimization problem \ref{Pe} will recover the components \[\b{L}\] and \[\b{S}\] exactly. To this end, we will show that such a \[\lambda_e\] belongs to a non-empty interval \[[\lambda_e^{\min}, \lambda_e^{\max}]\], where \[\lambda_e^{\min}\] and \[\lambda_e^{\max}\] are defined as
\begin{align}\label{eqn:lam}
\lambda_{e}^{\min} := \tfrac{1 + C_e}{1-C_e} ~\xi_e ~\text{and}~\lambda_e^{\max} := 
\tfrac{\sqrt{\alpha_\ell} (1-\mu) -\sqrt{r \alpha_u} \mu}{\sqrt{s_e}}.
\end{align}
Here, \[C_e(\alpha_u, \alpha_\ell,  \gamma_{\b{U}},  \gamma_{\b{V}}, s_e, d, k, \mu)\] where \[0 \leq C_e < 1\] is a constant that captures the relationship between different model parameters, and is defined as
\begin{align*}
C_e := \tfrac{c}{\alpha_\ell(1 - \mu)^2 - c},
\end{align*}
where \[c = \tfrac{ \alpha_u}{2}((1 + 2 \gamma_{\b{U}} )(\min(s_e, d)  + s_e \gamma_{\b{V}} ) +2 \gamma_{\b{V}}\min(s_e, nm)) - \tfrac{\alpha_\ell}{2}(\min(s_e, d)  + s_e \gamma_{\b{V}} )\]. 
Given these definitions, we have the following result for the entry-wise sparsity structure.
\begin{theorem} \label{theorem_entry}
Suppose \[\b{M} = \b{L} + \b{DS}\], where $\rk(\b{L})=r$ and \[\b{S}\] has at most \[s_e\] non-zeros, i.e., \[\|\b{S}\|_0 \leq s_e \leq s_e^{\max} := \tfrac{(1 - \mu)^2}{2}\tfrac{nm}{r}\], and the dictionary \[\b{D} \in \mathbb{R}^{f \times d}\] for \[d\leq f\] obeys the generalized frame property \eqref{frame} with frame bounds \[[\alpha_\ell, \alpha_u ]\], where \[0<\alpha_\ell \leq \tfrac{1}{(1 - \mu)^2}\], and \[ \gamma_{\b{U}}\] follows 
\begin{align}\label{A2}
\gamma_{\b{U}} \leq
\begin{cases}
\tfrac{(1 - \mu)^2 - 2s_e \gamma_{\b{V}}}{2s_e( 1 +  \gamma_{\b{V}})}, \text{ for }     s_e \leq \min ~(d, s_e^{\max})\\
\tfrac{(1 - \mu)^2 - 2s_e \gamma_{\b{V}}}{2(d + s_e \gamma_{\b{V}})}, \text{ for }   d<s_e \leq s_e^{\max}.
\end{cases}
\end{align}

Then given \[\mu \in [0, 1]\], \[ \gamma_{\b{U}}\] and \[ \gamma_{\b{V}} \in [r/nm, 1]\], and \[\xi_e\] defined in \eqref{eqn:mu}, \eqref{eqn:gamma}, \eqref{eqn:xi}, respectively, \[\lambda_e \in [\lambda_e^{\min}, \lambda_e^{\max} ]\] with \[\lambda_e^{\max} >\lambda_e^{\min}\geq 0\] defined in \eqref{eqn:lam}, solving \ref{Pe} will recover matrices \[\b{L}\] and \[\b{S}\].
\end{theorem}
We observe that the conditions for the recovery of \[(\b{L, S})\] are closely related to the incoherence measures (\[\mu\], \[ \gamma_{\b{V}}\], and \[ \gamma_{\b{U}}\]) between the low-rank part, \[\b{L}\], the dictionary, \[\b{D}\], and the sparse component \[ \b{S}\]. In general, smaller sparsity, rank, and incoherence parameters are sufficient for ensuring the recovery of the components for a particular problem. This is in line with our intuition that the more distinct the two components, the easier it should be to tease them apart. For our HS demixing problem, this indicates that a target of interest can be localized as long as its the spectral signature is appropriately different from the other materials in the scene. 

\vspace*{-5pt}
\subsection{Recovery for Column-wise Sparsity Case}\label{sec:rpca_c}

For the column-wise sparsity model, recall that any pair in the oracle model described in \textcolor{blue}{\textbf{D.\ref{def:oracle}}} is considered optimal. To this end, we first establish the sufficient conditions for the existence of such an optimal pair  \[(\b{L}_0, \b{S}_0)\] by the following lemma.
\begin{lemma}\label{lem:unique}
Given \[\b{M}\], \[\b{D}\], and \[\left( \c{L}, {\c{S}_c}, \c{D} \right)\], any pair \[( \b{L}_0, \b{S}_0 ) \in \{ \b{M}, \c{U}, \c{I}_{\c{S}_c}  \}\] satisfies \[\spann\{\col(\b{L}_0) \} = \c{U}\] and \[\csupp(\b{S}_0) = \c{I}_{\c{S}_c} \] if \[\mu < 1\].
\end{lemma}
In essence, we need the incoherence parameter \[\mu\] to be strictly smaller than \[1\]. Next, analogous to the entry-wise case, we show that \[\lambda_c\] belongs to a non-empty interval \[ [\lambda_c^{\min}, \lambda_c^{\max}]\], using which solving \ref{Pc} recovers an optimal pair in the oracle model \textcolor{blue}{\textbf{D.\ref{def:oracle}}} in accordance with Lemma~\ref{lem:unique}. Here, for a constant $C_c := \tfrac{ \alpha_u}{\alpha_\ell}\tfrac{1}{{(1 - \mu)^2}}\gamma_{\b{V}}\beta_{\b{U}}$,  \[\lambda_c^{\min}\] and \[\lambda_c^{\max}\] are defined as 
\begin{align}\label{eqn:lamb_c}
\hspace{-0.05in}\lambda_c^{\min} := \tfrac{\xi_c + \sqrt{r s_c \alpha_u}\mu C_c}{1 - s_cC_c}~\text{and}~
\lambda_c^{\max} := \tfrac{\sqrt{\alpha_{\ell}} (1 - \mu) - \sqrt{r \alpha_{u}}\mu}{\sqrt{s_c}}.
\end{align}
This leads us to the following result for the column-wise case.

\begin{theorem}\label{theorem_col}
	Suppose \[\b{M} = \b{L} + \b{D}\b{S}\] with \[(\b{L}, \b{S})\] defining the oracle model \[\{ \b{M}, \c{U}, \c{I}_{\c{S}_c} \}\], where \[\rk(\b{L})=r\], \[|\c{I}_{\c{S}_c}|=s_c\] for \[s_c \leq s_c^{\max} := \tfrac{\alpha_\ell}{ \alpha_u\gamma_{\b{V}}}\cdot\tfrac{(1 - \mu)^2}{\beta_{\b{U}}} \]. Given \[\mu \in [0,1)\], \[\beta_{\b{U}}\], \[\gamma_{\b{V}} \in [r/nm, 1]\], \[\xi_c\] as defined in \eqref{eqn:mu}, \eqref{eqn:beta}, \eqref{eqn:gamma}, \eqref{eqn:xi}, respectively, and any \[\lambda_c \in [\lambda_c^{\min}, \lambda_c^{\max}]\], for \[\lambda_c^{\max} > \lambda_c^{\min} \geq 0\] defined in \eqref{eqn:lamb_c}, solving \ref{Pc} will recover a pair of components \[(\b{L}_0, \b{S}_0)  \in \{ \b{M}, \c{U}, \c{I}_{\c{S}_c}  \}\], if the dictionary \[\b{D} \in \mathbb{R}^{f \times d}\] obeys the generalized frame property \textcolor{blue}{\textbf{D}.}\ref{frame} with frame bounds \[[\alpha_\ell, \alpha_u ]\], for \[\alpha_\ell>0\].	
\end{theorem}

Theorem~\ref{theorem_col} outlines the sufficient conditions under which the solution to the optimization problem \ref{Pc} will be in the oracle model defined in \textcolor{blue}{\textbf{D.\ref{def:oracle}}}. 
Here, for a case where \[1 \lesssim \alpha_{l} \leq \alpha_{u} \lesssim 1\], which can be easily met by a tight frame when \[f >d\], constant \[\tfrac{(1-\mu)^2}{\beta_U}\], and \[\gamma_{\b{V}} = \Theta(\tfrac{r}{nm})\], we have  \[s_c^{\max} = \c{O}(\tfrac{nm}{r})\], which is of same order as in the Outlier Pursuit (OP) \cite{Xu2010}. Moreover, our numerical results in \cite{Rambhatla18LrTheo} show that \ref{Pc} can be much more robust than OP, and may recover \[\{ \c{U}, \c{I}_{\b{C}} \}\] even when the rank of \[\b{L}\] is high and the number of outliers \[s_c\] is a constant proportion of \[m\]. This implies that, \ref{Pc} will succeed as long as the dictionary \[\b{D}\] can successfully represent the target of interest while rejecting the columns of the data matrix \[\b{M}\] corresponding to materials other than the target. 

\begin{algorithm}[!t]
	\caption{APG Algorithm for \ref{Pe} and \ref{Pc}, adapted from \cite{Mardani2012}}
	\label{algo}
	\begin{algorithmic}
		\REQUIRE $\b{M}$, $\b{D}$, $\lambda$, $v$, $\nu_0$, $\bar{\nu}$, and $L_f = \lambda_{max}\left([\b{I} ~\b{D}]^\top [\b{I} ~\b{D}] \right)$   
		\STATE \hspace{-1.25em} \textbf{Initialize:} $\b{L}[0] =  \b{L}[-1] = \textbf{0}_{L\times T}$, $\b{S}[0] = \b{S}[-1] = \textbf{0}_{F\times T}$, $t[0] = t[-1] = 1$, and set $k = 0$.    
		\hspace{-3em} \WHILE {not converged}
		\vspace{6pt}
		\STATE Generate points \[\b{T}_L[k]\] and \[\b{T}_S[k]\] using momentum:
		\vspace{3pt}
		\STATE ~~~~~~$\b{T}_L[k] = \b{L}[k] + \frac{t[k-1]-1}{t[k]}(\b{L}[k] - \b{L}[k-1])$,
		\STATE ~~~~~~$\b{T}_S[k] = \b{S}[k] + \frac{t[k-1]-1}{t[k]}(\b{S}[k] - \b{S}[k-1])$.
		\vspace{6pt}
		\STATE Take a gradient step using these points :
		\vspace{3pt}
		\STATE ~~~~~~$\b{G}_{L}[k] = \b{T}_{L}[k] + \frac{1}{L_f}(\b{M} - \b{T}_L[k]-\b{D}\b{T}_S[k])$,
		\STATE ~~~~~~$\b{G}_{S}[k] = \b{T}_{S}[k] + \frac{1}{L_f}\b{D}^\top(\b{M} - \b{T}_L[k]-\b{D}\b{T}_S[k])$.
		\vspace{6pt}
		\STATE Update Low-rank part via singular value thresholding:
		\vspace{3pt}
		\STATE ~~~~~~$\b{U} \b{\Sigma} \b{V}^\top = \text{svd}(\b{G}_L[k])$,
		\STATE ~~~~~~$\b{L}[k+1] = \b{U} \mathcal{S}_{\nu[k]/L_f}(\b{\Sigma})\b{V}^\top$.
		\vspace{6pt}
		\STATE Update the Dictionary Sparse part:
		\vspace{3pt}
		\STATE ~~~$\b{S}[k+1] = \begin{cases}
		\mathcal{S}_{\nu[k]\lambda_e/L_f}(\b{G}_S[k]), &\text{for \ref{Pe}},\\
		\mathcal{C}_{\nu[k]\lambda_c/L_f}(\b{G}_S[k]), &\text{for \ref{Pc}}.\end{cases}$
		\vspace{6pt}
		\STATE Update the momentum term parameter \[t[k+1]\]:
		\vspace{3pt}
		\STATE ~~~~~~$t[k+1] = \tfrac{1+\sqrt{4t^2[k]+1}}{2}$.
		\vspace{6pt}
		\STATE Update the continuation parameter \[\nu[k+1]\]:
		\vspace{3pt}
		\STATE ~~~~~~$\nu[k+1] = \max\{v\nu[k],\bar{\nu}\}$.
		\vspace{6pt}	 
		\STATE $k$ $\leftarrow$ $k + 1$   
		\hspace{-3em} \ENDWHILE
		\STATE \hspace{-1.25em} \textbf{return} $\b{L}[k]$, $\b{S}[k]$
		\end{algorithmic}
		\end{algorithm}
		\vspace*{-5pt}
\section{Algorithmic Considerations}
\label{sec:optimization}
The optimization problems of interest, \ref{Pe} and \ref{Pc}, for the entry-wise and column-wise case, respectively, are convex but non-smooth. To solve for the components of interest, we adopt the accelerated proximal gradient (APG) algorithm, as shown in Algorithm~\ref{algo}. Note that \cite{Mardani2012} also applied the APG algorithm for \ref{Pe}, and we present a unified algorithm for both sparsity cases for completeness.
\vspace*{-5pt}
\subsection{Background}
The APG algorithm is motivated from a long line of work starting with \cite{Nesterov83}, which showed the existence of a first order algorithm with a convergence rate of \[\c{O}(1/k^2)\] for a smooth convex objective, where \[k\] denotes the iterations. Following this, \cite{Beck09} developed the popular  fast iterative shrinkage-thresholding algorithm (FISTA) which achieves this convergence rate for convex non-smooth objectives by accelerating the proximal gradient descent algorithm using a \textit{momentum term} (the term \[\tfrac{t[k-1]-1}{t[k]}\] in Algorithm~\ref{algo}) as prescribed by \cite{Nesterov83}. As a result, it became a staple to solve a wide range of convex non-smooth tasks including matrix completion \cite{Toh10}, and robust PCA \cite{Chen09} and its variants \cite{Mardani2012, Xu2010}. Also, recently \cite{Karimi16} has shown further improvements in the rate of convergence.

In addition to the momentum term, the APG procedure operates by evaluating the gradient at a point further in the direction pointed by the negative gradient. Along with faster convergence, this insight about the next point minimizes the oscillations around the optimum point;
see \cite{Beck09} and references therein. 

\vspace*{-5pt}
\subsection{Discussion of Algorithm~\ref{algo}}

For the optimization problem of interest, we solve an unconstrained problem by transforming the equality constraint to a least-square term which penalizes the fit. In particular, the problems of interest we will solve via the APG algorithm are given by 
\begin{align}
\underset{\b{L}, \b{S}}{\min~} \nu\|\b{L}\|_* + \nu\lambda_e \|\b{S}\|_1 + \tfrac{1}{2} \|\b{M} - \b{L} - \b{DS}\|_{\rm F}^2 \label{Pe_apg}
\end{align}
for the entry-wise sparsity case, and
\begin{align}
\underset{\b{L}, \b{S}}{\min~} \nu\|\b{L}\|_* + \nu\lambda_c \|\b{S}\|_{1,2} + \tfrac{1}{2}\|\b{M} - \b{L} - \b{DS}\|_{\rm F}^2, \label{Pc_apg}
\end{align}
for the column-wise sparsity case.
We note that although for the application at hand, the thin dictionary case with ($f\geq d$) might be more useful in practice, Algorithm~\ref{algo} allows for the use of fat dictionaries ($f<d$) as well. 

\input{table_ip_ew}

Algorithm~\ref{algo} also employs a continuation technique \cite{Chen09}, which can be viewed as a ``warm start'' procedure. Here, we initialize the parameter \[\nu_0\] at some large value and geometrically reduced until it reaches a value \[\bar{\nu}\]. A smaller choice of \[\bar{\nu}\] results in a solution which is closer to the optimal solution of the constrained problem. Further, as \[\nu\] approaches zero, \eqref{Pe_apg} and \eqref{Pc_apg} recover the optimal solution of \ref{Pe} and \ref{Pc}, respectively. Moreover, Algorithm~\ref{algo} also utilizes the knowledge of the smoothness constant \[L_f\] (the Lipschitz constant of gradient) to set the step-size parameter. 

Specifically, the APG algorithm requires that the gradient of the smooth part,
\begin{align*}
f(\b{L},\b{S}) := \tfrac{1}{2}\|\b{M} - \b{L} - \b{DS}\|_{\rm F}^2 = \tfrac{1}{2}\|\b{M} - \begin{bmatrix}
\b{I} & \b{D}
\end{bmatrix}
\begin{bmatrix}
\b{L}\\\b{S}
\end{bmatrix}\|_{\rm F}^2
\end{align*}
of the convex objectives shown in \eqref{Pe_apg} and \eqref{Pc_apg} is Lipschitz continuous with minimum Lipschitz constant \[L_f\]. Now, since the gradient \[\nabla f(\b{L},\b{S})\] with respect to \[\begin{bmatrix}
\b{L} &\b{S}
\end{bmatrix}^\top \]is given by 
\begin{align*}
\nabla f(\b{L},\b{S}) =\begin{bmatrix}
\b{I} & \b{D}
\end{bmatrix}^\top(\b{M} - \begin{bmatrix}
\b{I} & \b{D}
\end{bmatrix}
\begin{bmatrix}
\b{L}\\\b{S}
\end{bmatrix}),
\end{align*}
we have that the gradient \[\nabla f\] is Lipschitz continuous as
\begin{align*}
\|\nabla f(\b{L}_1,\b{S}_1) - \nabla f(\b{L}_2,\b{S}_2) \| \leq L_f \|\begin{bmatrix}
\b{L}_1\\\b{S}_1
\end{bmatrix} - \begin{bmatrix}
\b{L}_2\\\b{S}_2
\end{bmatrix}\|,
\end{align*}
where 
\begin{align*}
L_f = \|\begin{bmatrix}
\b{I} & \b{D}
\end{bmatrix}^\top\begin{bmatrix}
\b{I} & \b{D}
\end{bmatrix}\|= \lambda_{\max} (\begin{bmatrix}
\b{I} & \b{D}
\end{bmatrix}^\top\begin{bmatrix}
\b{I} & \b{D}
\end{bmatrix}),
\end{align*}
as shown in Algorithm~\ref{algo}.

The update of the low-rank component  and the sparse matrix \[\b{S}\] for the entry-wise case, both involve a soft thresholding step, \[\mathcal{S}_{\tau}(.)\], where for a matrix \[\b{Y}\], \[\mathcal{S}_{\tau}(\b{Y}_{ij})\] is defined as 
\begin{align*}
\mathcal{S}_{\tau}(\b{Y}_{ij}) = {\rm sgn}{(\b{Y}_{ij})}\max({|\b{Y}_{ij} - \tau|, 0}).
\end{align*}
In case of the low-rank part we apply this function to the singular values (therefore referred to as \textit{singular value thresholding}) \cite{Toh10}, while for the update of the dictionary sparse component, we apply it to the sparse coefficient matrix \[\b{S}\].

The low-rank update step for the column-wise case remains the same as for the entry-wise case. However, for the update of the column-wise case we threshold the columns of \[\b{S}\] based on their column norms, i.e., for a column \[\b{Y}_j\] of a matrix \[\b{Y}\], the column-norm based soft-thresholding function, \[\mathcal{C}_{\tau}(.)\] is defined as 
\begin{align*}
\mathcal{C}_{\tau}(\b{Y}_{j}) = \max({\b{Y}_{j} - \tau{\b{Y}_{j}}/{\|\b{Y}_{j}\|}}).
\end{align*}

\vspace*{-5pt}
\subsection{Parameter Selection}

Since the choice of regularization parameters by our main theoretical results contain quantities (such as incoherence etc.) that cannot be evaluated in practice, we employ a grid-search strategy over the range of admissible values for the low-rank and dictionary sparse component to find the best values of the regularization parameters. We now discuss the specifics of the grid-search for each sparsity case. 


\subsubsection{Selecting parameters for the entry-wise case}\label{sec:opt_ew_param}

The choice of parameters \[\nu\] and \[\lambda_e\] in Algorithm~\ref{algo} is based on the optimality conditions of the optimization problem shown in \eqref{Pe_apg}. As presented in \cite{Mardani2012}, the range of parameters \[\nu\] and \[\nu\lambda_e\] associated with the low-rank part \[\b{L}\] and the sparse coefficient matrix \[\b{S}\], respectively, lie in \[\nu \in \{0,\|\b{M}\|\}\] and \[\nu\lambda_e \in \{0,\|\b{D}^\top\b{M}\|_\infty\}\], i.e., for Algorithm~\ref{algo} \[\nu_0 = \|\b{M}\|\].

These ranges for \[\nu\] and \[\nu\lambda_e\] are derived using the optimization problem shown in \eqref{Pe_apg}. Specifically, we find the largest values of these regularization parameters which yield a \[(\b{0}, \b{0})\] solution for the pair \[(\b{L}_0, \b{S}_0)\] by analyzing the optimality conditions of \eqref{Pe_apg}. This value of the regularization parameter then defines the upper bound on the range. For instance, let \[\lambda_*:=\nu\] and \[\lambda_1 := \nu\lambda_e\], then the optimality condition is given by
\begin{align*}
\lambda_*\partial_\b{L} \|\b{L}\|_*   - (\b{M} - \b{L} - \b{DS}) = 0,
\end{align*}
where the sub-differential set \[\partial_\b{L} \|\b{L}\|_* \] is defined as
\begin{align*}
\partial_\b{L} \|\b{L}\|_* \Bigr|_{\b{L} = \b{L}_0} = \{ \b{UV^\top} + \b{W} : \|\b{W}\| \leq 1, \c{P}_{\c{L}} (\b{W}) = \b{0} \}.	
\end{align*}
Therefore, for a zero solution pair \[(\b{L}_0, \b{S}_0)\] we have that
\begin{align*}
\{ \lambda_*\b{W} =  \b{M}: \|\b{W}\| \leq 1, \c{P}_{\c{L}} (\b{W}) = \b{0} \},
\end{align*}
which yields the condition that \[\|\b{M}\| \leq \lambda_*\]. Therefore, the maximum value of \[\lambda_*\] which drives the low-rank part to an all-zero solution is \[\|\b{M}\|\]. 

Similarly, for the dictionary sparse component the optimality condition for choosing \[\lambda_1\] is given by
\begin{align*}
\lambda_1 \partial_\b{S} \|\b{S}\|_1  - \b{D}^\top (\b{M} - \b{L} - \b{DS}) = 0,
\end{align*}
where the the sub-differential set \[\partial_\b{S} \|\b{S}\|_1\] is defined as
\begin{align*}
\partial_\b{S} \|\b{S}\|_1 \Bigr|_{\b{S} = \b{S}_0}  = \{ \text{sign}(\b{S}_0) + \b{F} : \|\b{F}\|_{\infty} \leq 1, \c{P}_{\c{S}_e} (\b{F}) = \b{0} \}.
\end{align*}
Again, for a zero solution pair \[(\b{L}_0, \b{S}_0)\] we need that
\begin{align*}
\{ \lambda_1\b{F} = \b{D}^\top \b{M} : \|\b{F}\|_{\infty} \leq 1, \c{P}_{\c{S}_e} (\b{F}) = \b{0} \},
\end{align*}
which implies that \[\|\b{D}^\top \b{M}\|_\infty \leq \lambda_1\]. Meaning, that the maximum value of \[\lambda_1\] that drives the dictionary sparse part to zero is \[\|\b{D}^\top \b{M}\|_\infty\].

\subsubsection{Selecting parameters for the column-wise case}

Again, the choice of parameters \[\nu\] and \[\lambda_c\] is derived from the optimization problem shown in \eqref{Pc_apg}. In this case, the range of parameters \[\nu\] and \[\nu\lambda_c\] associated with the low-rank part \[\b{L}\] and the sparse coefficient matrix \[\b{S}\], respectively, lie in \[\nu \in \{0,\|\b{M}\|\}\] and \[\nu\lambda_e \in \{0,\|\b{D}^\top\b{M}\|_{\infty,2}\}\], i.e., for Algorithm~\ref{algo} \[\nu_0 = \|\b{M}\|\]. The range of regularization parameters are evaluated using the analysis similar to the entry-wise case, by analyzing the optimality conditions for the optimization problem shown in \eqref{Pc_apg}, instead of \eqref{Pe_apg}.

\vspace*{-5pt}
\section{Experimental Evaluation}
\label{sec:exp}
We now evaluate the performance of the proposed technique on real HS data. We begin by introducing the dataset used for the simulations, following which we describe the experimental set-up and present the results.
\vspace*{-5pt}
\subsection{Data}
\noindent\textbf{Indian Pines Dataset}:
We first consider the ``Indian Pines'' dataset \cite{HSdat}, which was collected over the Indian Pines test site in North-western Indiana in the June of 1992 using the Airborne Visible/Infrared Imaging Spectrometer (AVIRIS) \cite{AVIRIS} sensor, a popular choice for collecting HS images for various remote sensing applications. This dataset consists of spectral reflectances across $224$ bands in wavelength of ranges $400-2500$ nm from a scene which is composed mostly of agricultural land along with two major dual lane highways, a rail line and some built structures, as shown in Fig.~\ref{fig:gt_data}(a). The dataset is further processed by removing the bands corresponding to those of water absorption, which results in a HS data-cube with dimensions $\{145 \times 145 \times 200\}$ is as visualized in Fig.~\ref{fig:data}. Here, $n = m = 145$ and $f = 200$. This modified dataset is available as ``corrected Indian Pines'' dataset \cite{HSdat}, with the ground-truth containing $16$ classes; Henceforth, referred to as the ``Indian Pines Dataset". We form the data matrix  $\b{M} \in \mathbb{R}^{f \times nm}$ by stacking each voxel of the image side-by-side, which results in a $\{200 \times 145^2\}$ data matrix $\b{M}$. We will analyze the performance of the proposed technique for the identification of the stone-steel towers (class $16$ in the dataset), shown in Fig.~\ref{fig:gt_data}(a), which constitutes about $93$ voxels in the dataset.  

\vspace{6pt}
\noindent\textbf{Pavia University Dataset}:
Acquired using Reflective Optics System Imaging Spectrometer (ROSIS) sensor, the Pavia University Dataset \cite{PaviaUSdat} consists of spectral reflectances across $103$ bands (in the range $430-860$ nm) of an urban landscape over northern Italy. The selected subset of the scene, a $\{201 \times 131 \times 103\}$ data-cube, mainly consists of buildings, roads, painted metal sheets and trees, as shown in Fig.~\ref{fig:gt_data}(b). Note that class-$3$ corresponding to ``Gravel'' is not present in the selected data-cube considered here. For our demixing task, we will analyze the localization of target class $5$, corresponding to the painted metal sheets, which constitutes $707$ voxels in the scene. Note that for this dataset $n = 201$, $m = 131$ and $f = 103$. 

\begin{figure}[!t]
	\centering
	\begin{tabular}{cc}
		\hspace{-4pt}\includegraphics[width=0.58\linewidth]{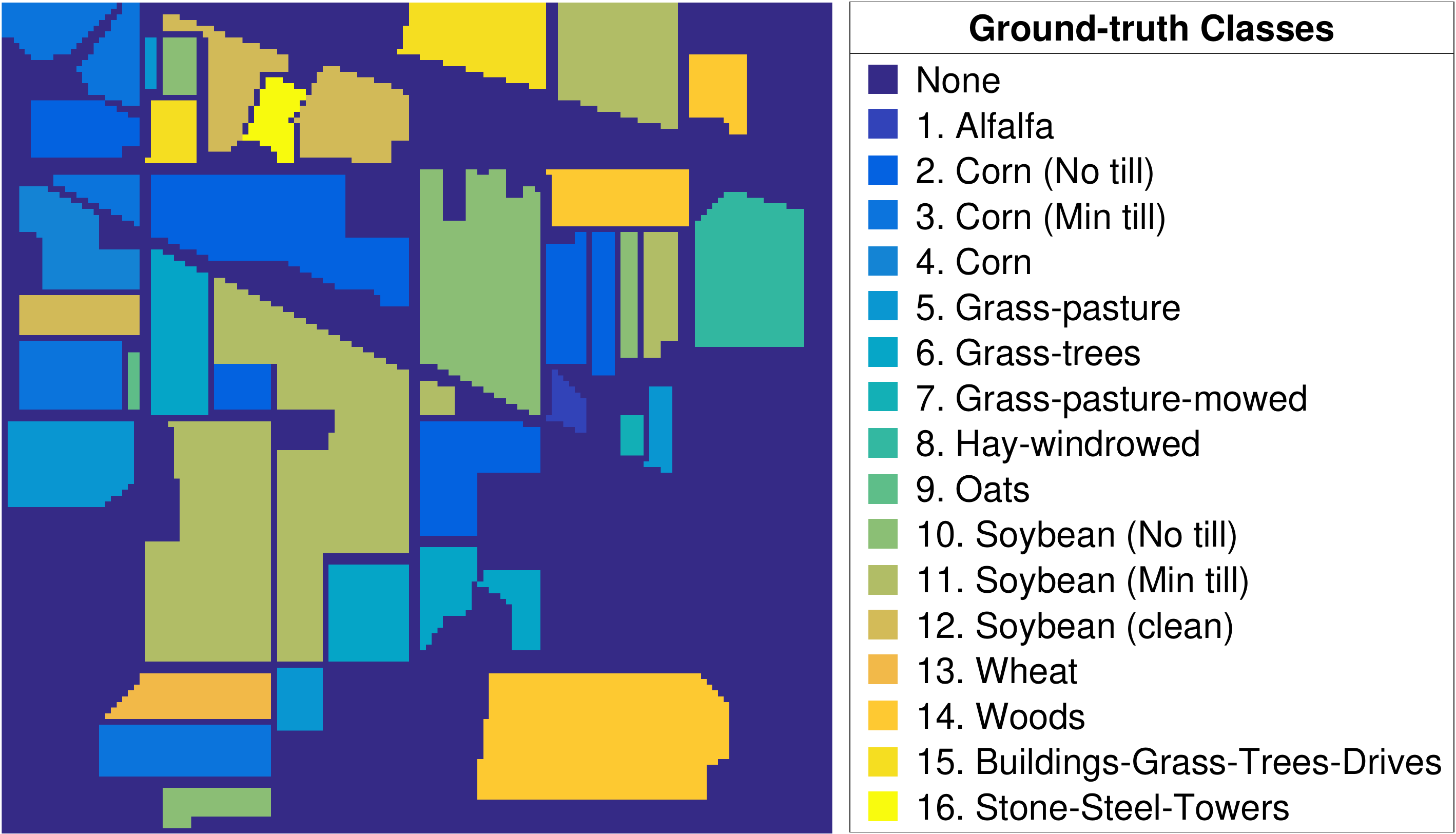}&\hspace{-14pt}
		\includegraphics[width=0.405\linewidth]{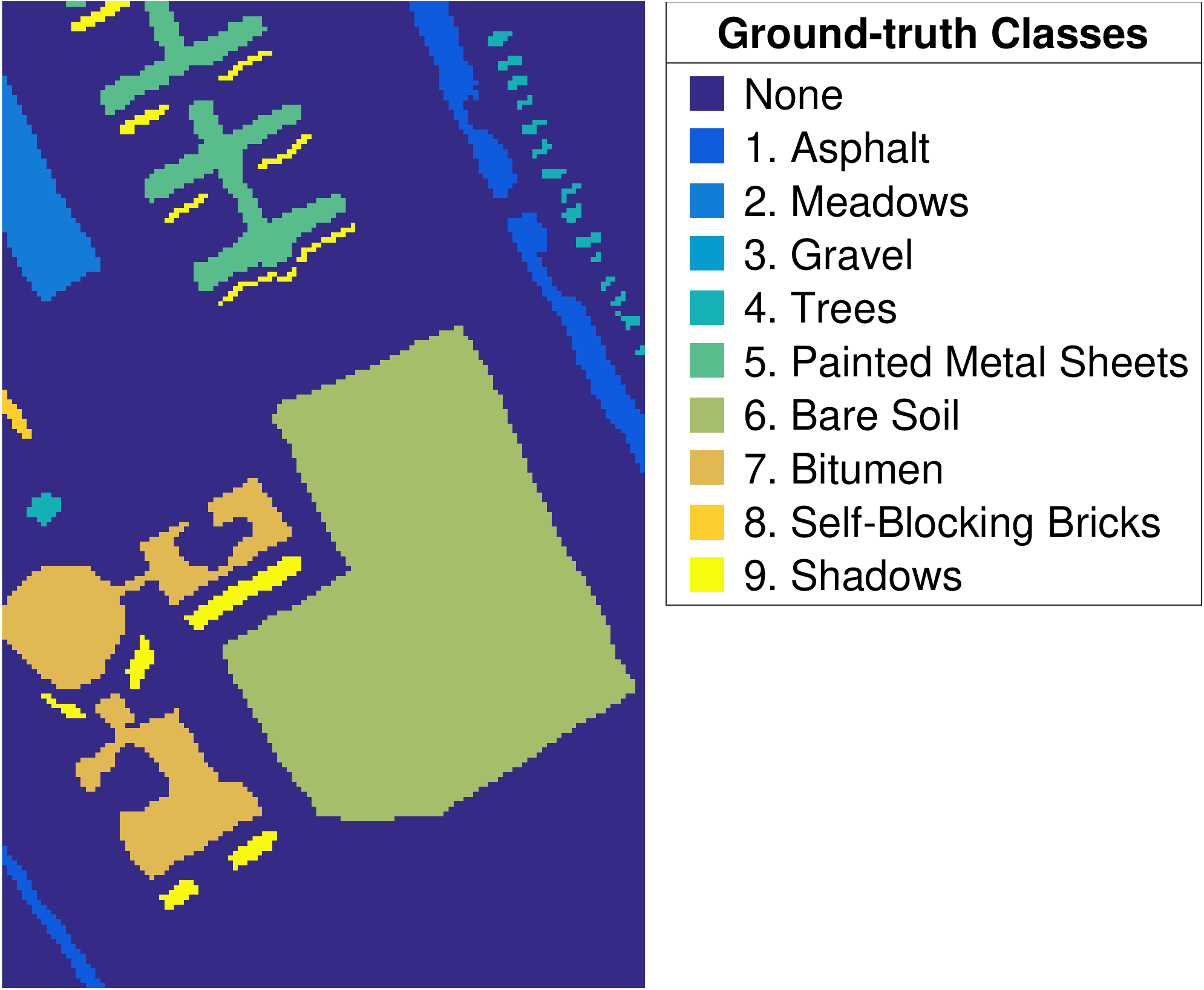}\\
		\hspace{-4pt}	(a) Indian Pines &\hspace{-14pt} (b) Pavia University\\
	\end{tabular}
	\caption{Ground-truth classes in the datasets. Panels (a) and (b) show the ground truth classes for the Indian Pines dataset \cite{HSdat} and Pavia University dataset \cite{PaviaUSdat}, respectively.} 
	\label{fig:gt_data} 
	\vspace{-6pt}
\end{figure}

\begin{algorithm}[!b]
	\caption{Dictionary Learning \cite{Mairal10, Lee2007}}
	\label{algo_dl}
	\begin{algorithmic}
		\REQUIRE Data $\b{Y} \in \RR^{f \times p}$, regularization parameter $\rho$, and the number of dictionary elements $d$.
	    \ENSURE The dictionary $\b{D} \in \RR^{f \times d}$
		\STATE \hspace{-1.25em} \textbf{Initialize:} $\hat{\b{A}} \leftarrow \b{0}_{d \times p}$, $\hat{\b{D}}$ with $\c{N}(0,1)$ entries and columns normalized to have norm $1$, $\hat{\b{Y}} = \hat{\b{D}}\hat{\b{A}}$, and tolerance $\epsilon$ .
		\hspace{-3em} \WHILE {$\tfrac{\|\b{Y} - \hat{\b{Y} \|_{\rm F}}}{\|\b{Y}\|_{\rm F}} \geq \epsilon$}
		\STATE Update Coefficient Matrix  $\b{A}$:
		\begin{align}\label{dl:coeff}
		\hat{\b{A}}= \underset{\b{A}}{\rm{arg. min}}  \|\b{Y} - \b{\hat{D}A}\|_{\rm F}^2 + \rho \|\b{A}\|_1
		\end{align}
		\vspace{-5pt}
		\STATE Update Dictionary $\b{D}$:
		\begin{align}\label{dl:dict}
		\hat{\b{D}} = \underset{\b{D}:\|\b{D}_i\|=1}{{\rm arg. min}} ~ \|\b{Y} - \b{DA}\|_{\rm F}^2
		\end{align}
		\vspace{-5pt}
		\STATE Form Estimate of Data $\hat{\b{Y}}$:
		\STATE ~~~~~~$\hat{\b{Y}} = \hat{\b{D}}\hat{\b{A}}$
		\hspace{-3em} \ENDWHILE
		\STATE \hspace{-1.25em} \textbf{return} $\hat{\b{D}}$
		\end{algorithmic}
\end{algorithm}
	   \input{table_pu_ew}   
\subsection{Dictionary}
We form the known dictionary \[\b{D}\] two ways: 1) where a (thin) dictionary is learned based on the voxels using Algorithm~\ref{algo_dl}, and 2) when the dictionary is formed by randomly sampling voxels from the target class. This is to emulate the ways in which we can arrive at the dictionary corresponding to a target -- 1) where the \textit{exact signatures} are not available, and/or there is noise, and 2) where we have access to the exact signatures of the target, respectively. Note that, the optimization procedures for \ref{Pe} and \ref{Pc} are agnostic to the selection of the dictionary.

In our experiments for case 1), we learn the dictionary using the target class data \[\b{Y}\in \RR^{f \times p} \] via Algorithm~\ref{algo_dl}, which (approximately) solves the following optimization problem,
\begin{align*}
\hat{\b{D}} = \underset{\b{D}:\|\b{D}_i\|=1, \b{A}}{{\rm arg. min}} ~ \|\b{Y} - \b{DA}\|_{\rm F}^2 + \rho \|\b{A}\|_1,
\end{align*}

Algorithm~\ref{algo_dl} operates by alternating between updating the sparse coefficients \eqref{dl:coeff} via FISTA \cite{Beck09} and dictionary \eqref{dl:dict} via the Newton method \cite{Wright06}.  

For case 2), the columns of the dictionary are set as the known data voxels of the target class. Specifically, instead of learning a dictionary based on a target class of interest, we set it as the exact signatures observed previously. Note that for this case, the dictionary is not normalized at this stage since the specific normalization depends on the particular demixing problem of interest, discussed shortly. In practice, we can store the un-normalized dictionary  \[\b{D}\] (formed from the voxels), consisting of actual \textit{signatures} of the target material, and can normalize it after the HS image has been acquired. 

%
%
 
 \input{table_ip_cw}
 \vspace*{-10pt}
\subsection{Experimental Setup}\label{sec:exp_setup}

\noindent\textbf{Normalization of data and the dictionary:} For normalizing the data, we divide each element of the data matrix $\b{M}$ by $\|\b{M}\|_{\infty}$ to preserve the inter-voxel scaling. For the dictionary, in the learned dictionary case, i.e., case 1), the dictionary already has unit-norm columns as a result of Algorithm~\ref{algo_dl}. Further, when the dictionary is formed from the data directly, i.e., for case 2), we divide each element of \[\b{D}\] by $\|\b{M}\|_{\infty}$, and then normalize the columns of \[\b{D}\], such that they are unit-norm.  

\vspace{6pt}
\noindent\textbf{Dictionary selection for the Indian Pines Dataset}: For the learned dictionary case, we evaluate the performance of the aforementioned techniques for both entry-wise and column-wise settings for two dictionary sizes, $d=4$ and $d=10$, for three values of the regularization parameter $\rho$, used for the initial dictionary learning step, i.e., $\rho = 0.01,~0.1$ and $0.5$. Here, the parameter $\rho$ controls the sparsity during the initial dictionary learning step; see Algorithm~\ref{algo_dl}.  For the case when dictionary is selected from the voxels directly, we randomly select $15$ voxels from the target class-$16$ to form our dictionary.

\vspace{6pt}
\noindent\textbf{Dictionary selection for the Pavia University Dataset}: Here, for the learned dictionary case, we evaluate the performance of the aforementioned techniques for both entry-wise and column-wise settings for a dictionary of size $d=30$ for three values of the regularization parameter $\rho$, used for the initial dictionary learning step, i.e., $\rho = 0.01,~0.1$ and $0.5$. Further, we randomly select $60$ voxels from the target class-$5$, when the dictionary is formed from the data voxels.

\vspace{6pt}
\noindent\textbf{Comparison with matched filtering (MF)-based approaches}: In addition to the robust PCA-based and OP-based techniques introduced in Section~\ref{rel_tech}, we also compare the performance of our techniques with two MF-based approaches. These MF-based techniques are agnostic to our model assumptions, i.e., entry-wise or column-wise sparsity cases. Therefore, the following description of these techniques applies to both sparsity cases.

For the first MF-based technique, referred to as \textcolor{blue}{MF}, we form the inner-product of the column-normalized data matrix $\b{M}$, denoted as $\b{M}_n$,  with the dictionary $\b{D}$, i.e., $\b{D^\top M}_n$, and select the maximum absolute inner-product per column. For the second MF-based technique, \textcolor{blue}{MF$^\dagger$}, we perform matched filtering on the pseudo-inversed data \[\b{\widetilde{M} = D^\dagger M}\]. Here, the matched filtering corresponds to finding maximum absolute entry for each column of the column-normalized $\b{\widetilde{\b{M}}}$. Next, in both cases we scan through $1000$ threshold values between $(0, 1]$ to generate the results. 

\vspace{6pt}
\noindent\textbf{Performance Metrics}: We evaluate the performance of these techniques via the receiver operating characteristic (ROC) plots. ROC plots are a staple for analysis of classification performance of a binary classifier in machine learning; see \cite{James2013} for details. Specifically, it is a plot between the true positive rate (TPR) and the false positive rate (FPR), where a higher TPR (close to $1$) and a lower FPR (close to $0$) indicate that the classifiier performs detects all the elements in the class while rejecting those outside the class. 

A natural metric to gauge good performance is the area under the curve (AUC) metric. It indicates the area under the ROC curve, which is maximized when TPR \[=1\] and FPR \[=0\], therefore, a higher AUC is preferred. Here, an AUC of $0.5$ indicates that the performance of the classifier is roughly as good as a coin flip. As a result, if a classifier has an AUC $<0.5$, one can improve the performance by simply inverting the result of the classifier. This effectively means that AUC is evaluated after ``flipping'' the ROC curve. In other words, this means that the classifier is good at rejecting the class of interest, and taking the complement of the classifier decision can be used to identify the class of interest.  

In our experiments, MF-based techniques often exhibit this phenomenon. Specifically, when the dictionary contains element(s) which resemble the average behavior of the spectral signatures, the inner-product between the normalized data columns and these dictionary elements may be higher as compared to other distinguishing dictionary elements. Since, MF-based techniques rely on the maximum inner-product between the normalized data columns and the dictionary, and further since the spectral signatures of even distinct classes are highly correlated; see, for instance Fig.~\ref{fig:data_corr}, where MF-based approaches in these cases can effectively reject the class of interest. This leads to an AUC \[<0.5\]. Therefore, as discussed above, we invert the result of the classifier (indicated as $(\cdot)_*$ in the tables) to report the best performance. If using MF-based techniques, this issue can potentially be resolved in practice by removing the dictionary elements which tend to resemble the average behavior of the spectral signatures.

\vspace*{-5pt}
\subsection{Parameter Setup for the Algorithms}

\noindent\textbf{Entry-wise sparsity case}: 
We evaluate and compare the performance of the proposed method \ref{Pe} with \ref{Prob_rpca} (described in Section~\ref{priorart}), \textcolor{blue}{MF}, and \textcolor{blue}{MF$^\dagger$}. Specifically, we evaluate the performance of these techniques via the receiver \edit{operating} characteristic (ROC) plot for the Indian Pines dataset and the Pavia University dataset, with the results shown in Table~\ref{res_tab}(a)-(d) and Table~\ref{res_tab_pu_ew}(a)-(c), respectively.


\input{table_pu_cw}
For the proposed technique, we employ the accelerated proximal gradient (APG) algorithm shown in Algorithm~\ref{algo} and discussed in Section~\ref{sec:optimization} to solve the optimization problem shown in \ref{Pe}. Similarly, for \ref{Prob_rpca} we employ the APG algorithm with transformed data matrix $\widetilde{\b{M}}$, while setting $\b{D = I}$. 

With reference to selection of tuning parameters for the APG solver for \eqref{Pe} (\ref{Prob_rpca}, respectively), we choose $v = 0.95$, $\nu =\|\mb{M}\|$ ($\nu =\|\b{\widetilde{M}}\|$), $\bar{\nu} = 10^{-4}$, and scan through $100$ values of $\lambda_e$ in the range $\lambda_e \in (0, {\|\b{D^\top M}\|_{\infty}}/{\|\b{M}\|} ]$ ($\lambda_e \in (0, {\|\b{\widetilde{M}}{\|_{\infty}}/{\|\b{\widetilde{M}}\|} }]$), to generate the ROCs. 
We threshold the resulting estimate of the sparse part $\b{S} \in \RR^{d \times nm}$ based on its column norm. We choose the threshold such that the AUC metric is maximized for both cases (\ref{Pe} and \ref{Prob_rpca}). 

\vspace{0.05in}
\noindent\textbf{Column-wise sparsity case}: For this case, we evaluate and compare the performance of the proposed method \ref{Pc} with \ref{Prob_op} (as described in Section~\ref{priorart}), \textcolor{blue}{MF}, and \textcolor{blue}{MF$^\dagger$}. The results for the Indian Pines dataset and the Pavia University dataset as shown in Table~\ref{res_tab_ip_cw}(a)-(d) and Table~\ref{res_tab_pu_cw}(a)-(c), respectively. 

As in the entry-wise sparsity case, we employ the accelerated proximal gradient (APG) algorithm presented in Algorithm~\ref{algo} to solve the optimization problem shown in \ref{Pc}. Similarly, for \ref{Prob_op} we employ the APG with transformed data matrix $\widetilde{\b{M}}$, while setting $\b{D = I}$. For the tuning parameters for the APG solver for \eqref{Pc} (\ref{Prob_op}, respectively), we choose $v = 0.95$, $\nu =\|\mb{M}\|$ ($\nu =\|\b{\widetilde{M}}\|$), $\bar{\nu} = 10^{-4}$, and scan through $100$ $\lambda_c$s in the range $\lambda_c \in (0, {\|\b{D^\top M}\|_{\infty,2}}/{\|\b{M}\|} ]$ ($\lambda_c \in (0, {\|\b{\widetilde{M}}{\|_{\infty,2}}/{\|\b{\widetilde{M}}\|} }]$), to generate the ROCs. 
As in the previous case, we threshold the resulting estimate of the sparse part $\b{S} \in \RR^{d \times nm}$ based on its column norm.



 \captionsetup{justification=justified}

\subsection{Analysis}
Table~\ref{res_tab}--\ref{res_tab_pu_ew} and Table~\ref{res_tab_ip_cw}--\ref{res_tab_pu_cw} show the ROC characteristics and the classification performance of the proposed techniques \ref{Pe} and \ref{Pc}, for two datasets under consideration, respectively, under various choices of the dictionary $\b{D}$ and regularization parameter $\rho$ for Algorithm~\ref{algo_dl}. We note that both proposed techniques \ref{Pe} and \ref{Pc} on an average outperform competing techniques, emerging as the most reliable techniques across different dictionary choices for the demixing task at hand; see Tables~\ref{res_tab}(d), \ref{res_tab_pu_ew}(c), \ref{res_tab_ip_cw}(d), and \ref{res_tab_pu_cw}(c). 

Further, the performance of \ref{Pc} is slightly better than \ref{Pe}. This can be attributed to the fact that the column-wise sparsity model does not require the columns of \[\b{S}\] to be sparse themselves. As alluded to in Section~\ref{sec:sp_choice}, this allows for higher flexibility in the choice of the dictionary elements for the thin dictionary case.

In addition, we see that the matched filtering-based techniques (and even \ref{Prob_op} based technique for $d=4 $ and $\rho=0.1$ in Table~\ref{res_tab_ip_cw}) exhibit ``flip'' or inversion of the ROC curve. As described in Section~\ref{sec:exp_setup}, this phenomenon is an indicator that a classifier is better at rejecting the target class. In case of MF-based technique, this is a result of a dictionary that contains an element that resembles the average behavior of the spectral responses. A similar phenomenon is at play in case of the \ref{Prob_op} for $d=4 $ and $\rho=0.1$ in Table~\ref{res_tab_ip_cw}. Specifically, here the inversion indicates that the dictionary is capable of representing the columns of the data \[\b{M}\] effectively, which leads to an increase in the corresponding column norms in their representation \[\widehat{\b{M}}\]. Coupled with the fact that the component \[\b{L}\] is no longer low-rank for this thin dictionary case (see our discussion in Section~\ref{rel_tech}), this results in rejection of the target class. On the other hand, our techniques \ref{Pe} and \ref{Pc} do not suffer from this issue. Moreover, note that across all the experiments, the thresholds for \ref{Prob_rpca} and \ref{Prob_op} are higher than their D-RPCA counterparts. This can also be attributed to the pre-multiplication by the pseudo-inverse of the dictionary \[\b{D}^\dagger\], which increases column norms based on the leading singular values of \[\b{D}\]. Therefore, using \ref{Pe}, when the target spectral response admits a sparse representation, and \ref{Pc}, otherwise, yield consistent and superior results as compared to related techniques considered in this work.


\begin{figure}[t]
 \centering
 \begin{tabular}{cP{0.01cm}cc}
 {\footnotesize \textbf{Data}}&&{\footnotesize$\b{L}$ }& {\footnotesize $\b{DS}$} \vspace{-2pt}\\
   \epsfig{file=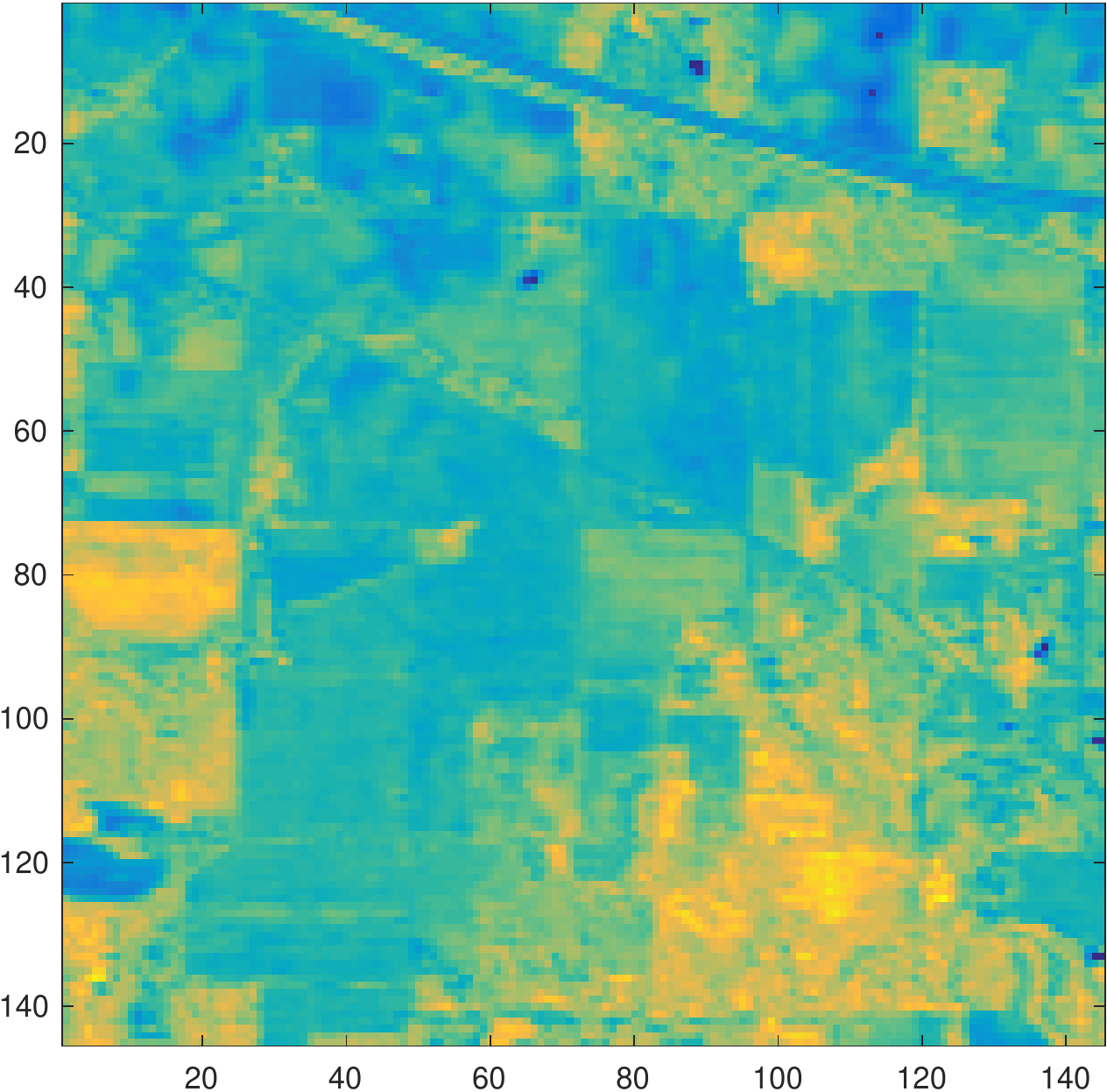,width=0.27\linewidth,clip=}&
    {\small\rotatebox{90}{ ~~~~~~~~~{Best $\lambda_e$}}} & 
    \epsfig{file=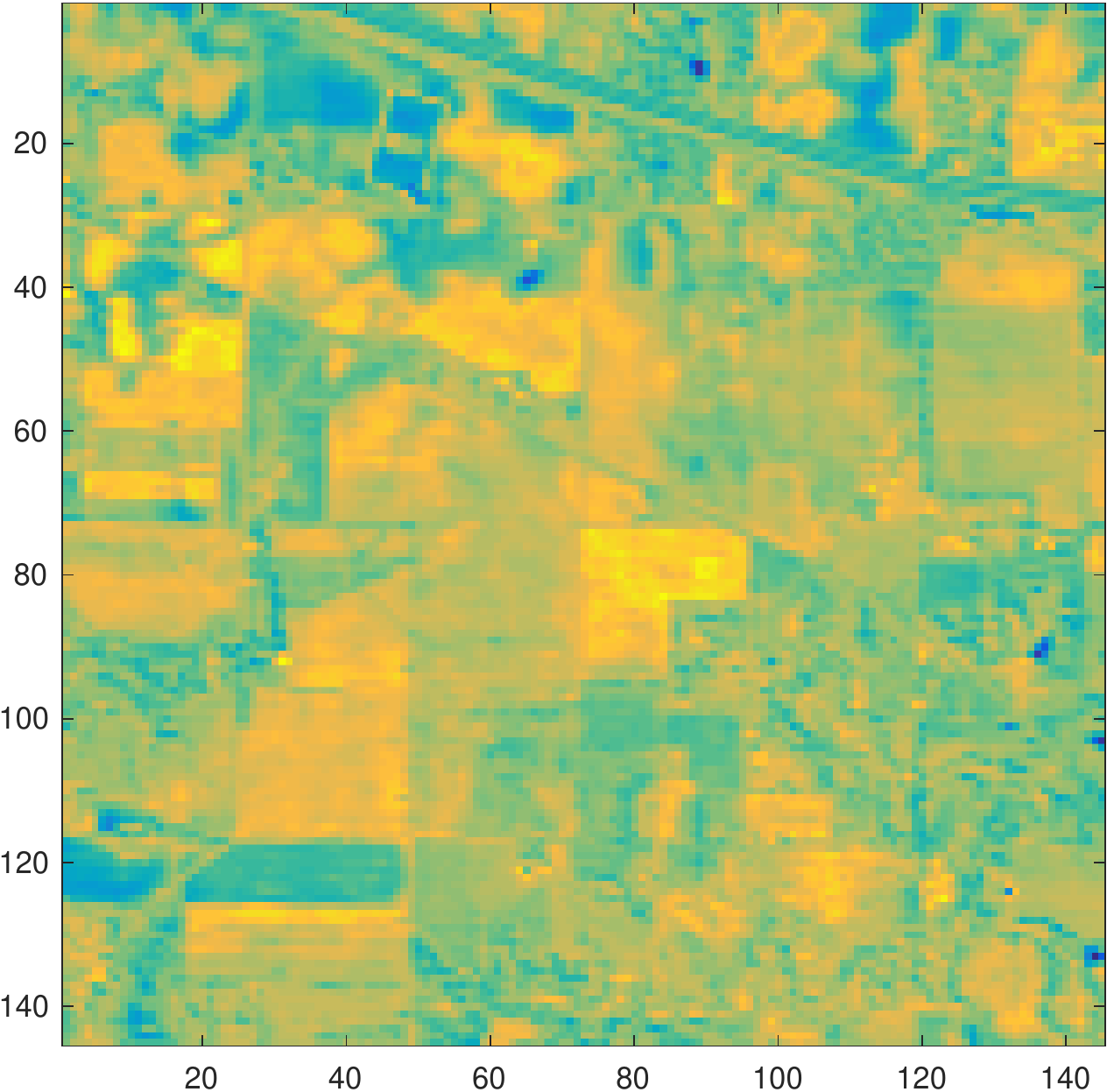,width=0.27\linewidth,clip=}&
    \epsfig{file=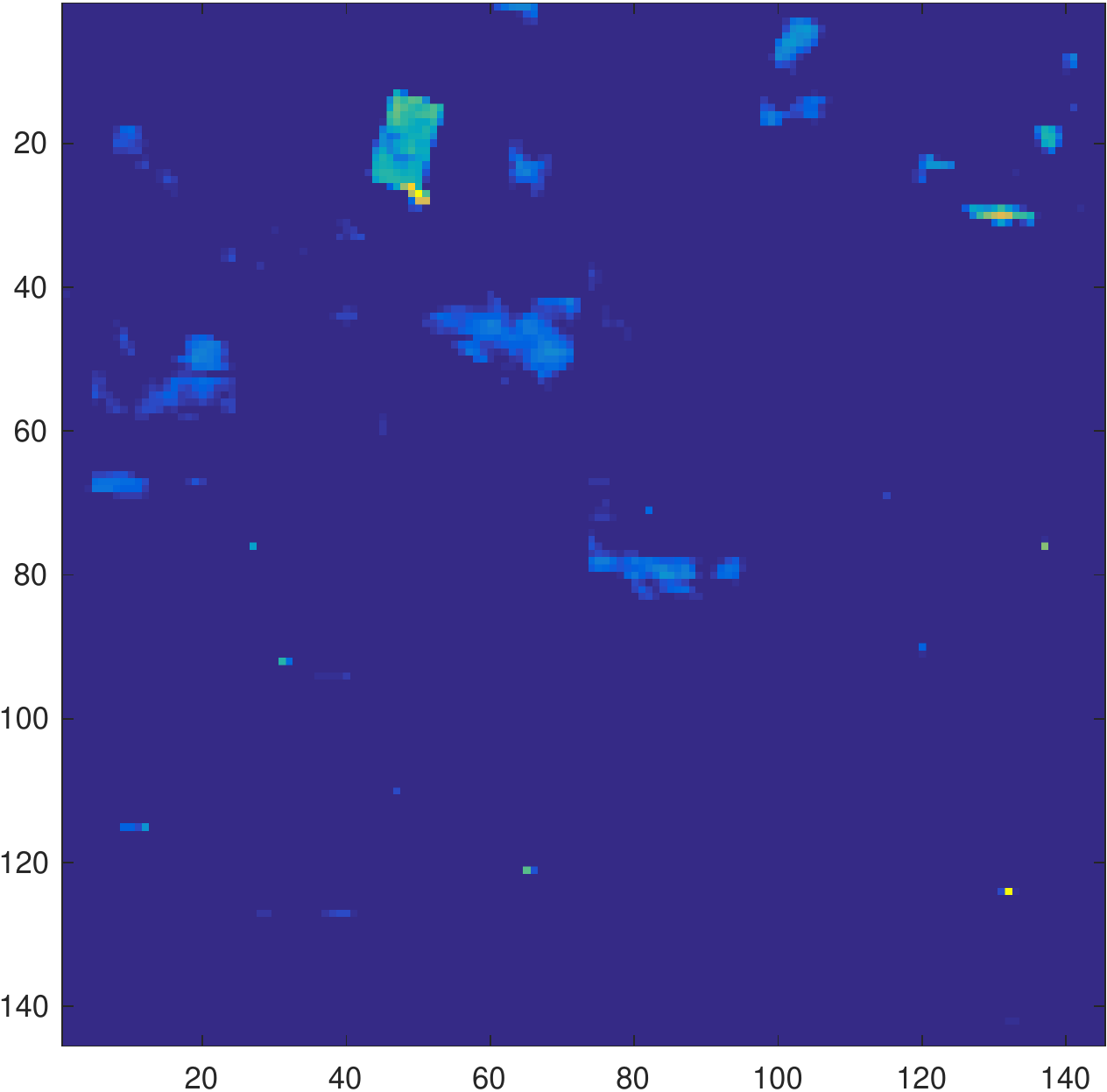,width=0.27\linewidth,clip=} \vspace{-5pt}\\ 
        (a) &&(c) & (d) \\
    \epsfig{file=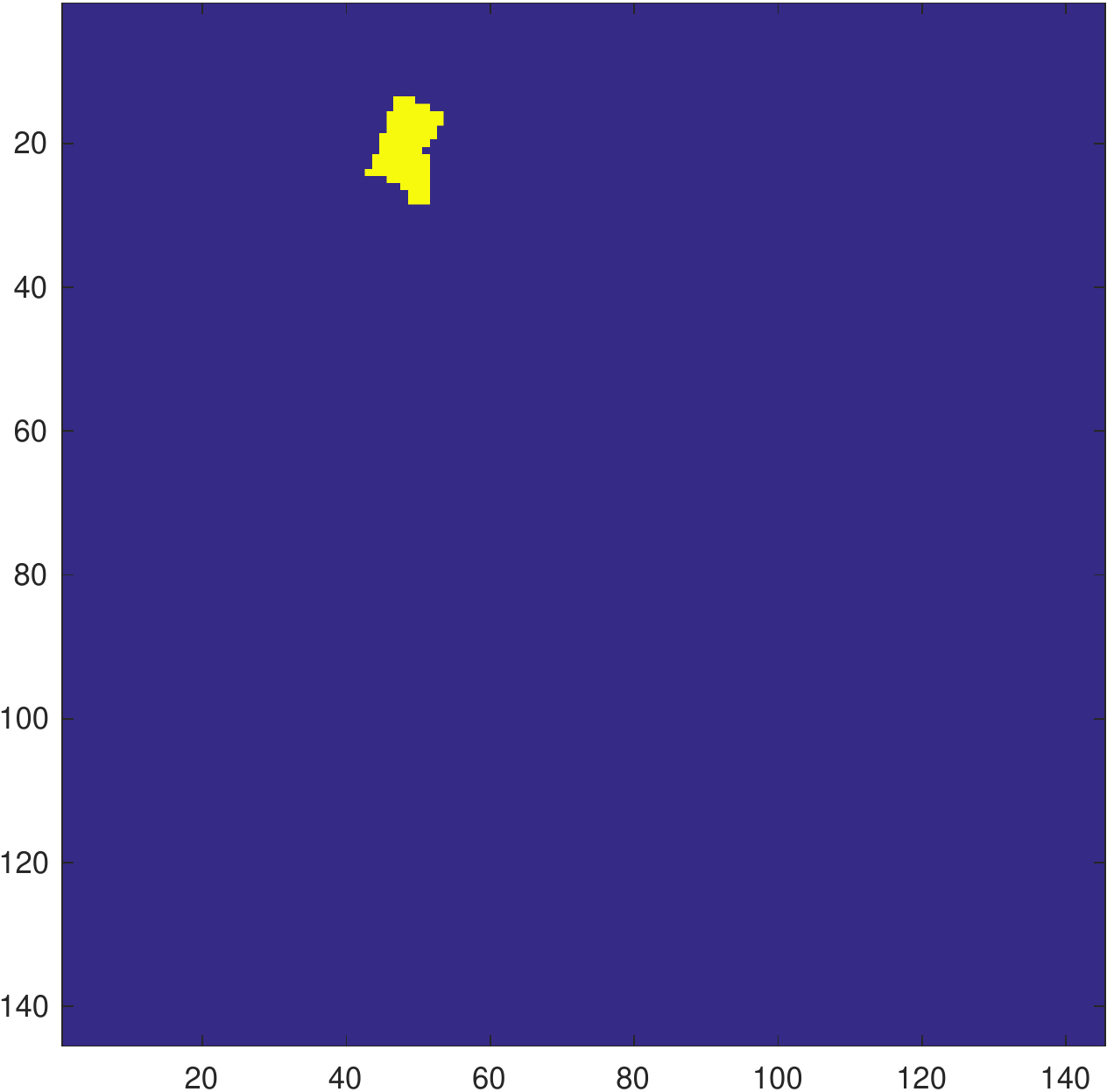,width=0.27\linewidth,clip=}&
    {\small\rotatebox{90}{ ~~~~{$85\%$ of $\lambda^{\max}_e$}}} &
    \epsfig{file=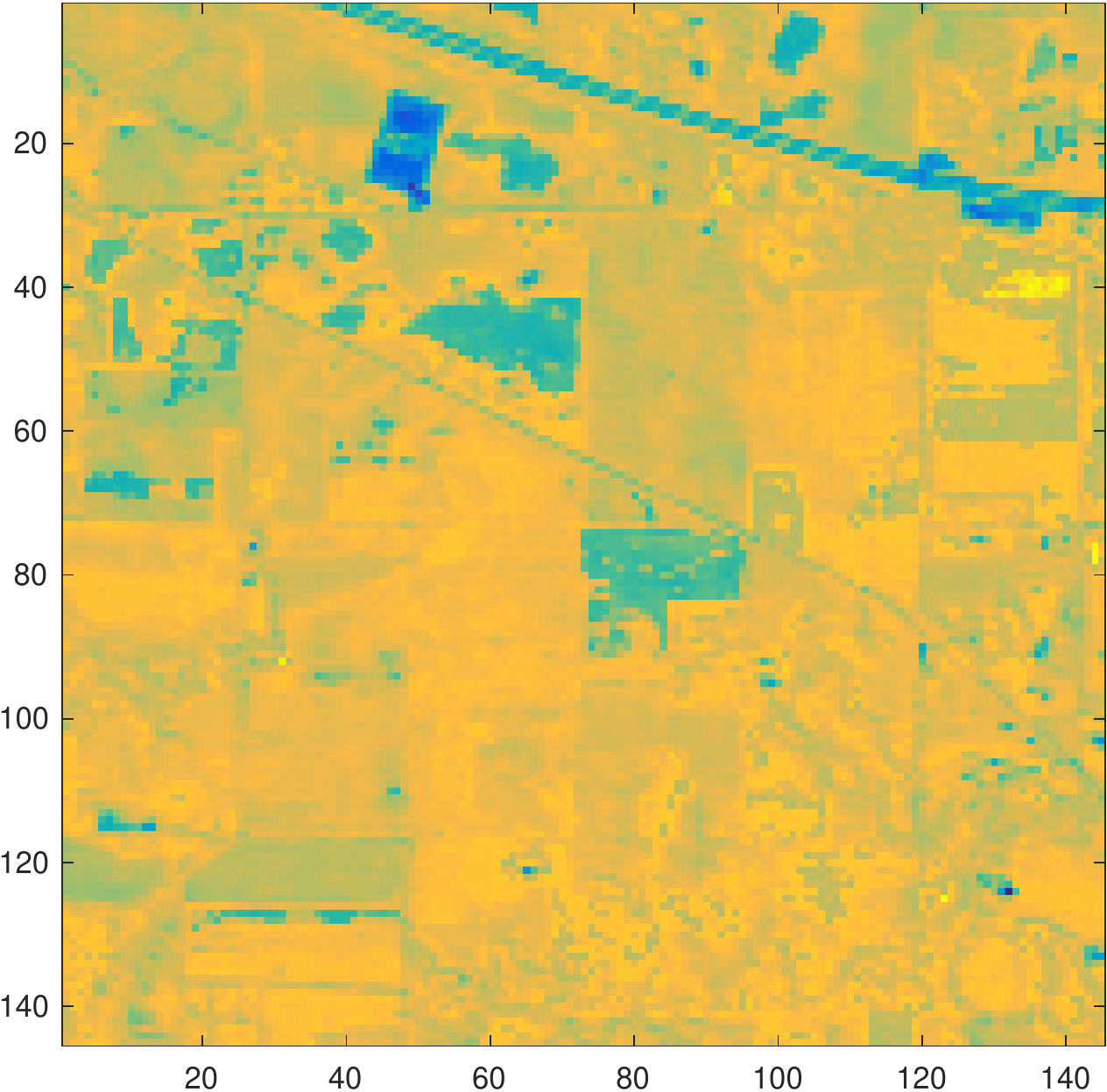,width=0.27\linewidth,clip=} &
    \epsfig{file=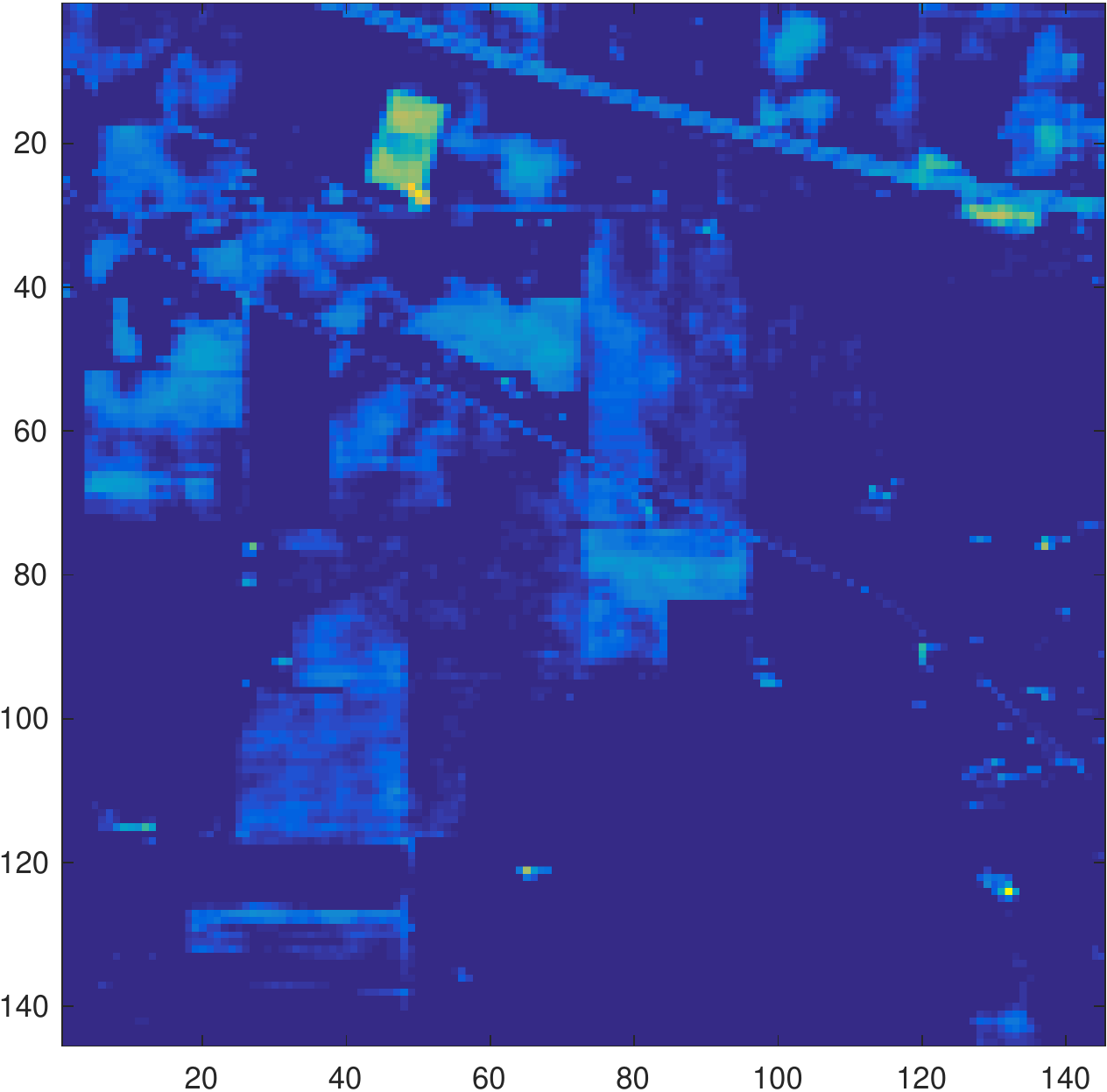,width=0.27\linewidth,clip=} \vspace{-5pt}\\ 
     (b)  && (e) & (f)
 \end{tabular}
 \caption{Recovery of the low-rank component \[\b{L}\] and the dictionary sparse component \[\b{DS}\] for different values of $\lambda$ for the proposed technique at { $f =50$-th} channel of the \cite{HSdat} (shown in panel (a)) corresponding to the results shown in Table~\ref{res_tab}(c). Panel (b) corresponds
 to the ground truth for class-$16$. Panel (c) and (d) show the recovery of the low-rank part and dictionary sparse part for a $\lambda$ at the best operating point. While, panels (e) and (f) show the recovery of these components at $\lambda_e = 85\%$ of $\lambda^{\max}_e$. Here, $\lambda^{\max}_e$ denotes the maximum value $\lambda_e$ can take; see Section~\ref{sec:opt_ew_param}.}\vspace*{-10pt}
 \label{figure:res_our}
 \end{figure}
 %
%

There are other interesting recovery results which warrant our attention. Fig.~\ref{figure:res_our} shows the low-rank and the dictionary sparse component recovered by \ref{Pe} for two different values of $\lambda_e$, for the case where we form the dictionary by randomly sampling the voxels (Table~\ref{res_tab}(c)) for the Indian Pines Dataset \cite{HSdat}.  Interestingly, we recover the rail tracks/roads running diagonally on the top-right corner, along with some low-density housing; see Fig~\ref{figure:res_our} (f). This is because the \textit{signatures} we seek (stone-steel towers) are similar to the signatures of the materials used in these structures. This further corroborates the applicability of the proposed approach in detecting the presence of a particular spectral \textit{signature} in a HS image. However, this also highlights potential drawback of this technique. As \ref{Pe}  and \ref{Pc} are based on identifying materials with similar composition, it may not be effective in distinguishing between very closely related classes, say two agricultural crops, also indicated by our theoretical results.
 
\section{Conclusions}
\label{sec:conclusion}
 We present a generalized robust PCA-based technique to localize a target in a HS image, based on the \textit{a priori} known spectral \textit{signature} of the material we wish to localize. We model the data as being composed of a low-rank component and a dictionary-sparse component, and consider two different sparsity patterns corresponding to different structural assumptions on the data, where the dictionary contains the \textit{a priori} known spectral \textit{signatures} of the target. We adapt the theoretical results of our previous work \cite{Rambhatla2016, Li2018, Rambhatla18LrTheo}, to present the conditions under which such decompositions recover the two components for the HS demixing task. Further, we evaluate and compare the performance of the proposed method via experimental evaluations for a classification task for different choices of the dictionary on real HS image datasets, and demostrate the applicability of the proposed techniques for a target localization in HS images.
	

\bibliographystyle{IEEEbib}
\bibliography{referLR}
\vspace{-0.6in}
\end{document}

%% file: commands.tex
\newcommand{ \mb}[1]{\mathbf{#1}}


%% file: table_ip_ew.tex
			\begin{table*}[!htbp]
			\caption{Entry-wise sparsity model for the Indian Pines Dataset. Simulation results are presented for our proposed approach \eqref{Pe}, robust-PCA based approach on transformed data \[\b{D^\dagger M}\] \eqref{Prob_rpca}, matched filtering (\textcolor{blue}{MF}) on original data \[\b{M}\], and matched filtering on transformed data \[\b{D^\dagger M}\] (\textcolor{blue}{MF$^\dagger$}), across dictionary elements $d$, and the regularization parameter for initial dictionary learning procedure $\rho$; see Algorithm~\ref{algo_dl} . Threshold selects columns with column-norm greater than threshold such that AUC is maximized.  For each case, the best performing metrics are reported in bold for readability. Further, $`` * "$ denotes the case where ROC curve was ``flipped'' (i.e. classifier output was inverted to achieve the best performance).}
			\label{res_tab}
		\captionsetup{justification=centering}
			\begin{subtable}{.5\linewidth}
			\captionsetup{font=small}
			 \centering
			 \caption{Learned dictionary, $d=4$}
		   	\scalebox{1}{
		   	\begin{tabular}{|P{0.5cm}|c|c|c|c|c|c|}
			\hline
			\multirow{2}{*}{\textbf{$d$}} & \multirow{2}{*}{$\rho$} & \multirow{2}{*}{\textbf{Method}}& \multirow{2}{*}{\textbf{Threshold }}&\multicolumn{2}{G|}{\textbf{Performance at best operating point} }& \multirow{2}{*}{\textbf{AUC}}\\ \cline{5-6}
			&&&&\textbf{TPR}&\textbf{FPR}&\\\hline
			\multirow{12}{*}{4} &	\multirow{4}{*}{0.01}
			  &\textbf{D-RPCA(E)}					&0.300&\textbf{0.979}&\textbf{0.023}&\textbf{0.989}\\\cline{3-7}
			&&\textbf{RPCA$^\dagger$} &0.650&0.957&0.049&0.974\\\cline{3-7}
			&&\textbf{MF$_*$}						&N/A&0.957&0.036&0.994\\\cline{3-7}
			&&\textbf{MF$_*^\dagger$}	   &N/A&0.914&0.104&0.946\\\cline{2-7}
			&	\multirow{4}{*}{0.1}
		  &\textbf{D-RPCA(E)}				    	&0.800&\textbf{0.989}&0.017&0.997\\\cline{3-7}
			&&\textbf{RPCA$^\dagger$} &0.800&0.989&0.014&0.997\\\cline{3-7}
			&&\textbf{MF}						&N/A&0.989&0.016&0.998\\\cline{3-7}
			&&\textbf{MF$^\dagger$}	   &N/A&0.989&\textbf{0.010}&\textbf{0.998}\\\cline{2-7}
			&	\multirow{4}{*}{0.5}
		  &\textbf{D-RPCA(E)}					    &0.600&\textbf{0.968}&\textbf{0.031}&\textbf{0.991}\\\cline{3-7}
			&&\textbf{RPCA$^\dagger$} &0.600&0.935&0.067&0.988\\\cline{3-7}
			&&\textbf{MF}						&N/A&0.548&0.474&0.555\\\cline{3-7}
			&&\textbf{MF$_*^\dagger$}	   &N/A&0.849&0.119&0.939\\\hline
			\end{tabular}
			\label{dl_4}
			}
			\end{subtable} 
			\begin{subtable}{.5\linewidth}
			\captionsetup{font=small}
			 \centering
			 \caption{Learned dictionary, $d=10$}
			   	\scalebox{1}{
			   	\begin{tabular}{|P{0.5cm}|c|c|c|c|c|c|}
				\hline
				\multirow{2}{*}{\textbf{$d$}} & \multirow{2}{*}{$\rho$} & \multirow{2}{*}{\textbf{Method}}& \multirow{2}{*}{\textbf{Threshold}}&\multicolumn{2}{G|}{\textbf{Performance at best operating point} }& \multirow{2}{*}{\textbf{AUC}}\\ \cline{5-6}
					&&&&\textbf{TPR}&\textbf{FPR}&\\\hline
				\multirow{12}{*}{10} &	\multirow{4}{*}{0.01}
				  &\textbf{D-RPCA(E)}					&0.600&0.935&0.060&0.972\\\cline{3-7}
				&&\textbf{RPCA$^\dagger$} &0.700&\textbf{0.978}&\textbf{0.023}&\textbf{0.990}\\\cline{3-7}
				&&\textbf{MF$_*$}						&N/A&0.624&0.415&0.681\\\cline{3-7}
				&&\textbf{MF$^\dagger_*$}	   &N/A&0.569&0.421&0.619\\\cline{2-7}
				&	\multirow{4}{*}{0.1}
			  &\textbf{D-RPCA(E)}					    &0.500&\textbf{0.968}&\textbf{0.029}&\textbf{0.993}\\\cline{3-7}
				&&\textbf{RPCA$^\dagger$} &0.500&0.871&0.144&0.961\\\cline{3-7}
				&&\textbf{MF$_*$}						&N/A&0.688&0.302&0.713\\\cline{3-7}
				&&\textbf{MF$^\dagger$}	   &N/A&0.527&0.469&0.523\\\cline{2-7}
				&	\multirow{4}{*}{0.5}
			  &\textbf{D-RPCA(E)}					    &1.000   &\textbf{0.978}&\textbf{0.031}&\textbf{0.996}\\\cline{3-7}
				&&\textbf{RPCA$^\dagger$} &2.200&0.849&0.113&0.908\\\cline{3-7}
				&&\textbf{MF}						&N/A&0.807&0.309&0.781\\\cline{3-7}
				&&\textbf{MF$^\dagger_*$}	   &N/A&0.527&0.465&0.539\\\hline
				\end{tabular}
				\label{dl_10}
				}
				\end{subtable}\vspace{10pt}
				\begin{subtable}{0.5\linewidth}
				\captionsetup{font=small}
				 \centering
				 \caption{Dictionary by sampling voxels, $d=15$}
					   	\scalebox{1}{
					   	\begin{tabular}{|P{0.5cm}|c|c|c|P{1.5cm}|P{0.9cm}|}
						\hline 
						\multirow{2}{*}{\textbf{$d$}}  & \multirow{2}{*}{\textbf{Method}}& \multirow{2}{*}{\textbf{Threshold}}&\multicolumn{2}{>{\bfseries\centering\arraybackslash} m{2.3cm}|}{\textbf{Performance at best operating point} }& \multirow{2}{*}{\textbf{AUC}}\\ \cline{4-5}
						&&&TPR&FPR&\\\hline
						\multirow{4}{*}{15}
						  &\textbf{D-RPCA(E)}					&0.300&\textbf{0.989}&\textbf{0.021}&\textbf{0.998}\\\cline{2-6}
						&\textbf{RPCA$^\dagger$} &3.000&0.849&0.146&0.900\\\cline{2-6}
						&\textbf{MF}						&N/A&0.957&0.085&0.978\\\cline{2-6}
						&\textbf{MF$^\dagger$}	   &N/A&0.796&0.217&0.857\\\hline
						\end{tabular}
						\label{orig_dict}
						}
						\end{subtable} 
							\begin{subtable}{.5\linewidth}
								\captionsetup{font=small}
								\centering
						\caption{Average performance}
						\scalebox{0.98}{	
						\begin{tabular}{|P{1.5cm}|c|c|c|c|c|c|}		
						\hline
						\multirow{2}{*}{\textbf{Method}}& \multicolumn{2}{G|}{\textbf{TPR}}& \multicolumn{2}{G|}{\textbf{FPR}}& \multicolumn{2}{G|}{\textbf{AUC}}\\ \cline{2-7}
						&Mean&St.Dev.&Mean&St.Dev.&Mean&St.Dev.\\ \hline
						\textbf{D-RPCA(E)}&	\textbf{0.972}&	\textbf{0.019}&	\textbf{0.030}&	\textbf{0.014}&	\textbf{0.991}&	\textbf{0.009}\\ \hline
						\textbf{RPCA$^\dagger$}&0.919&0.061&0.079&0.055&0.959&0.040\\ \hline
						\textbf{MF}&0.796&0.179&0.234&0.187&0.814&0.178\\ \hline
						\textbf{MF$^\dagger$}	&0.739&0.195&0.258&0.192&0.775&0.207\\ \hline
						\end{tabular}
						\label{overall_perf}
						}
						\end{subtable}
			\captionsetup{font=small}
			\vspace{-2pt}
			\end{table*}

%% file: table_pu_ew.tex
\begin{table*}[!htbp]
		 			\caption{Entry-wise sparsity model and Pavia University Dataset. Simulation results are presented for the proposed approach (\ref{Pe}), robust-PCA based approach on transformed data  (\ref{Prob_rpca}), matched filtering (\textcolor{blue}{MF}) on original data $\b{M}$, and matched filtering on transformed data $\b{D^\dagger M}$ (\textcolor{blue}{MF$^\dagger$}), across dictionary elements $d$, and the regularization parameter for initial dictionary learning step $\rho$. Threshold selects columns with column-norm greater than threshold such that AUC is maximized. For each case, the best performing metrics are reported in bold for readability. Further, $`` * "$ denotes the case where ROC curve was ``flipped'' (i.e. classifier output was inverted to achieve the best performance). }
		 			\label{res_tab_pu_ew}
		 			\captionsetup{justification=centering}
		 			\begin{subtable}{.5\linewidth}
		 				\captionsetup{font=footnotesize}
		 				\centering
		 				\caption{Learned dictionary, $d=30$}
		 				\scalebox{1}{
		 					\begin{tabular}{|P{0.5cm}|c|c|c|c|c|c|}
		 						\hline
		 						\multirow{2}{*}{\textbf{$d$}} & \multirow{2}{*}{$\rho$} & \multirow{2}{*}{\textbf{Method}}& \multirow{2}{*}{\textbf{Threshold }}&\multicolumn{2}{G|}{\textbf{Performance at best operating point} }& \multirow{2}{*}{\textbf{AUC}}\\ \cline{5-6}
		 						&&&&\textbf{TPR}&\textbf{FPR}&\\\hline
		 						\multirow{12}{*}{30} &	\multirow{4}{*}{0.01}
		 						&\textbf{D-RPCA(E)}					&0.150&\textbf{0.989}&\textbf{0.015}&\textbf{0.992}\\\cline{3-7}
		 						&&\textbf{RPCA$^\dagger$} &0.700&0.849&0.146&0.925\\\cline{3-7}
		 						&&\textbf{MF}						&N/A&0.929&0.073&0.962\\\cline{3-7}
		 						&&\textbf{MF$^\dagger$}	   &N/A&0.502&0.498&0.498\\\cline{2-7}
		 						&	\multirow{4}{*}{0.1}
		 						&\textbf{D-RPCA(E)}				    	&0.050&\textbf{0.982}&\textbf{0.019}&\textbf{0.992}\\\cline{3-7}
		 						&&\textbf{RPCA$^\dagger$} &3.000&0.638&0.374&0.664\\\cline{3-7}
		 						&&\textbf{MF}						&N/A&0.979&0.053&0.986\\\cline{3-7}
		 						&&\textbf{MF$^\dagger$}	   &N/A&0.620&0.381&0.660\\\cline{2-7}
		 						&	\multirow{4}{*}{0.5}
		 						&\textbf{D-RPCA(E)}					    &0.080&\textbf{0.982}&\textbf{0.019}&\textbf{0.992}\\\cline{3-7}
		 						&&\textbf{RPCA$^\dagger$} &2.500&0.635&0.381&0.671\\\cline{3-7}
		 						&&\textbf{MF}						&N/A&0.980&0.159&0.993\\\cline{3-7}
		 						&&\textbf{MF$_*^\dagger$}	   &N/A&0.555&0.447&0.442\\\hline
		 					\end{tabular}
		 					\label{dl_30:pu_ew}
		 				}
		 			\end{subtable} 
		 			\begin{subtable}{.5\linewidth}
		 				\captionsetup{font=footnotesize}
		 				\centering
		 				\caption{Dictionary by sampling voxels, $d=60$}
		 				\scalebox{1}{
		 					\begin{tabular}{|P{0.5cm}|c|c|c|c|c|}
		 						\hline 
		 						\multirow{2}{*}{\textbf{$d$}}  & \multirow{2}{*}{\textbf{Method}}& \multirow{2}{*}{\textbf{Threshold}}&\multicolumn{2}{G|}{\textbf{Performance at best operating point} }& \multirow{2}{*}{\textbf{AUC}}\\ \cline{4-5}
		 						&&&\textbf{TPR}&\textbf{FPR}&\\\hline
		 						\multirow{4}{*}{60}
		 						&\textbf{D-RPCA(E)}					&0.060&\textbf{0.986}&0.016&\textbf{0.995}\\\cline{2-6}
		 						&\textbf{RPCA$^\dagger$} &1.000&0.799&0.279&0.793\\\cline{2-6}
		 						&\textbf{MF}						&N/A&0.980&\textbf{0.011}&0.994\\\cline{2-6}
		 						&\textbf{MF$^\dagger$}	   &N/A&0.644&0.355&0.700\\\hline
		 					\end{tabular}
		 					\label{orig_dict:pu_ew}
		 				}
		 				\vspace{0.17cm}
		 				\caption{Average performance}
		 				\vspace{-3px}
		 				\scalebox{0.96}{	
		 					\begin{tabular}{|P{1.5cm}|c|c|c|c|c|c|}		
		 						\hline
		 						\multirow{2}{*}{\textbf{Method}}& \multicolumn{2}{G|}{\textbf{TPR}}& \multicolumn{2}{G|}{\textbf{FPR}}& \multicolumn{2}{G|}{\textbf{AUC}}\\ \cline{2-7}
		 						&\textbf{Mean}&\textbf{St.Dev.}&\textbf{Mean}&\textbf{St.Dev.}&\textbf{Mean}&\textbf{St.Dev.}\\ \hline
		 						\textbf{D-RPCA(E)}&	\textbf{0.984}&	\textbf{0.003}&	\textbf{0.014}&	\textbf{0.002}&	\textbf{0.993}&	\textbf{0.001}\\ \hline
		 						\textbf{RPCA$^\dagger$}&0.730&0.110&0.295&0.110&0.763&0.123\\ \hline
		 						\textbf{MF}&0.967&0.025&0.074&0.062&0.983&0.0149\\ \hline
		 						\textbf{MF$^\dagger$}	&0.580&0.064&0.420&0.065&0.575&0.125\\ \hline
		 					\end{tabular}
		 					\label{overall_perf:pu_ew}
		 				}
		 			\end{subtable}
		 			\captionsetup{font=small}
		 			\vspace{-12pt}
		 		\end{table*}

%% file: table_ip_cw.tex
	\begin{table*}[!bhtp]
				\caption{Column-wise sparsity model and Indian Pines Dataset. Simulation results are presented for the proposed approach (\ref{Pc}), Outlier Pursuit (OP) based approach on transformed data (\ref{Prob_op}), matched filtering (\textcolor{blue}{MF}) on original data $\b{M}$, and matched filtering on transformed data $\b{D^\dagger M}$ (\textcolor{blue}{MF$^\dagger$}), across dictionary elements $d$, and the regularization parameter for initial dictionary learning step $\rho$. Threshold selects columns with column-norm greater than threshold such that AUC is maximized. For each case, the best performing metrics are reported in bold for readability. Further, $`` * "$ denotes the case where ROC curve was ``flipped'' (i.e. classifier output was inverted to achieve the best performance).}
				
				\label{res_tab_ip_cw}
				\captionsetup{justification=centering}
				\begin{subtable}{0.5\linewidth}
					\captionsetup{font=footnotesize}
					\centering
					\caption{Learned dictionary, $d=4$}
					\scalebox{1}{
						\begin{tabular}{|P{0.5cm}|c|c|c|c|c|c|}
							\hline
							\multirow{2}{*}{\textbf{$d$}} & \multirow{2}{*}{$\rho$} & \multirow{2}{*}{\textbf{Method}}& \multirow{2}{*}{\textbf{Threshold }}&\multicolumn{2}{G|}{\textbf{Performance at best operating point} }& \multirow{2}{*}{\textbf{AUC}}\\ \cline{5-6}
							&&&&\textbf{TPR}&\textbf{FPR}&\\\hline
							\multirow{12}{*}{4} &	\multirow{4}{*}{0.01}
							&\textbf{D-RPCA(C)}					&0.905&\textbf{0.989}&\textbf{0.014}&\textbf{0.998}\\\cline{3-7}
							&&\textbf{OP$^\dagger$}        &0.895&0.989&0.015&0.998\\\cline{3-7}
							&&\textbf{MF$_*$}                   &N/A&0.656&0.376&0.611\\\cline{3-7}
							&&\textbf{MF$_*^\dagger$}    &N/A&0.624&0.373&0.639\\\cline{2-7}
							&	\multirow{4}{*}{0.1}
							&\textbf{D-RPCA(C)}	              &0.805&\textbf{0.989}&\textbf{0.013}&\textbf{0.998}\\\cline{3-7}
							&&\textbf{OP$^\dagger_*$}      &1.100&0.720&0.349&0.682\\\cline{3-7}
							&&\textbf{MF$_*$}                 &N/A&0.742&0.256&0.780\\\cline{3-7}
							&&\textbf{MF$^\dagger$}     &N/A&0.828&0.173&0.905\\\cline{2-7}
							&	\multirow{4}{*}{0.5}
							&\textbf{D-RPCA(C)}				&1.800&\textbf{0.989}&\textbf{0.010}&\textbf{0.998}\\\cline{3-7}
							&&\textbf{OP$^\dagger$}    &1.300&0.989&0.012&0.998\\\cline{3-7}
							&&\textbf{MF}                      &N/A&0.548&0.474&0.556\\\cline{3-7}
							&&\textbf{MF$_*^\dagger$}&N/A&0.849&0.146&0.939\\\hline
						\end{tabular}
						\label{dl_4_ip_cw}
					}
				\end{subtable} 
				\begin{subtable}{0.5\linewidth}
					\captionsetup{font=footnotesize}
					\centering
					\caption{Learned dictionary, $d=10$}
					\scalebox{1}{
						\begin{tabular}{|P{0.5cm}|c|c|c|c|c|c|}
							\hline
							\multirow{2}{*}{\textbf{$d$}} & \multirow{2}{*}{$\rho$} & \multirow{2}{*}{\textbf{Method}}& \multirow{2}{*}{\textbf{Threshold}}&\multicolumn{2}{G|}{\textbf{Performance at best operating point} }& \multirow{2}{*}{\textbf{AUC}}\\ \cline{5-6}
							&&&&\textbf{TPR}&\textbf{FPR}&\\\hline
							\multirow{12}{*}{10} &	\multirow{4}{*}{0.01}
							&\textbf{D-RPCA(C)}					&0.800&\textbf{0.946}&\textbf{0.016}&\textbf{0.993}\\\cline{3-7}
							&&\textbf{OP$^\dagger$}         &1.300&0.946&0.060&0.988\\\cline{3-7}
							&&\textbf{MF$_*$}						&N/A&0.946&0.060&0.987\\\cline{3-7}
							&&\textbf{MF$^\dagger_*$}	   &N/A&0.527&0.468&0.511\\\cline{2-7}
							&	\multirow{4}{*}{0.1}
							&\textbf{D-RPCA(C)}					    &0.550&\textbf{0.979}&\textbf{0.029}&\textbf{0.997}\\\cline{3-7}
							&&\textbf{OP$^\dagger$}             &0.800&0.893&0.112&0.928\\\cline{3-7}
							&&\textbf{MF$_*$}						&N/A&0.688&0.302&0.714\\\cline{3-7}
							&&\textbf{MF$^\dagger$}	          &N/A&0.527&0.470&0.523\\\cline{2-7}
							&	\multirow{4}{*}{0.5}
							&\textbf{D-RPCA(C)}					    &1.400  &\textbf{0.989}&\textbf{0.037}&\textbf{0.997}\\\cline{3-7}
							&&\textbf{OP$^\dagger$}             &0.800&0.807&0.148&0.847\\\cline{3-7}
							&&\textbf{MF}						        &N/A&0.807&0.309&0.781\\\cline{3-7}
							&&\textbf{MF$^\dagger_*$}	      &N/A&0.527&0.468&0.539\\\hline
						\end{tabular}
						\label{dl_10_ip_cw}
					}
				\end{subtable}\vspace{10pt}\\
				\hspace{-2pt}
				\begin{subtable}{0.5\linewidth}
					\captionsetup{font=small}
					\caption{Dictionary by sampling voxels, $d=15$}
					\scalebox{1}{
						   	\begin{tabular}{|P{0.5cm}|c|c|c|P{1.5cm}|P{0.9cm}|}
							\hline 
							\multirow{2}{*}{\textbf{$d$}}  & \multirow{2}{*}{\textbf{Method}}& \multirow{2}{*}{\textbf{Threshold}}&\multicolumn{2}{>{\bfseries\centering\arraybackslash} m{2.3cm}|}{\textbf{Performance at best operating point} }& \multirow{2}{*}{\textbf{AUC}}\\ \cline{4-5}
							&&&TPR&FPR&\\\hline
							\multirow{4}{*}{15}
							&\textbf{D-RPCA(C)}					&0.800&0.989&0.018&0.998\\\cline{2-6}
							&\textbf{OP$^\dagger$}           &2.200&0.882&0.126&0.900\\\cline{2-6}
							&\textbf{MF}						     &N/A&0.957&0.085&0.978\\\cline{2-6}
							&\textbf{MF$^\dagger$}	        &N/A&0.796&0.217&0.857\\\hline
						\end{tabular}
						\label{orig_dict_ip_cw}
					}
					\end{subtable}
					\begin{subtable}{0.5\linewidth}
					\captionsetup{font=small}
					\caption{Average performance}
					\scalebox{0.98}{	
						\begin{tabular}{|P{1.5cm}|c|c|c|c|c|c|}		
							\hline
							\multirow{2}{*}{\textbf{Method}}& \multicolumn{2}{G|}{\textbf{TPR}}& \multicolumn{2}{G|}{\textbf{FPR}}& \multicolumn{2}{G|}{\textbf{AUC}}\\ \cline{2-7}
							&Mean&St.Dev.&Mean&St.Dev.&Mean&St.Dev.\\ \hline
							\textbf{D-RPCA(C)}&	\textbf{0.981}&	\textbf{0.016}&	\textbf{0.020}&	\textbf{0.010}&	\textbf{0.997}&	\textbf{0.002}\\ \hline
							\textbf{OP$^\dagger$}&0.889&0.099&0.117&0.115&0.906&0.114\\ \hline
							\textbf{MF}                  &0.763&0.151&0.266&0.149&0.772&0.166\\ \hline
							\textbf{MF$^\dagger$}	&0.668&0.151&0.331&0.148&0.702&0.192\\ \hline
						\end{tabular}
						\label{overall_perf_ip_cw}
					}\vspace{16pt}
				\end{subtable}
				\captionsetup{font=small}
				\vspace*{-10pt}
			\end{table*}

%% file: table_pu_cw.tex
			\begin{table*}[!bhtp]
				\caption{Column-wise sparsity model and Pavia University Dataset. Simulation results for the proposed approach (\ref{Pc}), Outlier Pursuit (OP) based approach (\ref{Prob_op}), matched filtering (\textcolor{blue}{MF}) on original data $\b{M}$, and matched filtering on transformed data $\b{D^\dagger M}$ (\textcolor{blue}{MF$^\dagger$}), across dictionary elements $d$, and the regularization parameter for initial dictionary learning step $\rho$. Threshold selects columns with column-norm greater than threshold such that AUC is maximized. For each case, the best performing metrics are reported in bold for readability.  Further, $`` * "$ denotes the case where ROC curve was ``flipped'' (i.e. classifier output was inverted to achieve the best performance). }
				
				\label{res_tab_pu_cw}
				\captionsetup{justification=centering}
				\begin{subtable}{.5\linewidth}
					\captionsetup{font=footnotesize}
					\centering
					\caption{Learned dictionary, $d=30$}
					\scalebox{1}{
						\begin{tabular}{|P{0.5cm}|c|c|c|c|c|c|}
							\hline
							\multirow{2}{*}{\textbf{$d$}} & \multirow{2}{*}{$\rho$} & \multirow{2}{*}{\textbf{Method}}& \multirow{2}{*}{\textbf{Threshold }}&\multicolumn{2}{G|}{\textbf{Performance at best operating point} }& \multirow{2}{*}{\textbf{AUC}}\\ \cline{5-6}
							&&&&\textbf{TPR}&\textbf{FPR}&\\\hline
							\multirow{12}{*}{30} &	\multirow{4}{*}{0.01}
							&\textbf{D-RPCA(C)}					&0.065&\textbf{0.990}&\textbf{0.015}&\textbf{0.991}\\\cline{3-7}
							&&\textbf{OP$^\dagger$} &0.800&0.7581&0.3473&0.705\\\cline{3-7}
							&&\textbf{MF}						&N/A&0.929&0.073&0.962\\\cline{3-7}
							&&\textbf{MF$^\dagger$}	   &N/A&0.502&0.50&0.498\\\cline{2-7}
							&	\multirow{4}{*}{0.1}
							&\textbf{D-RPCA(C)}				    	&0.070&\textbf{0.996}&\textbf{0.022}&\textbf{0.994}\\\cline{3-7}
							&&\textbf{OP$^\dagger$} &0.100&0.989&0.3312&0.904\\\cline{3-7}
							&&\textbf{MF}						&N/A&0.979&0.053&0.986\\\cline{3-7}
							&&\textbf{MF$^\dagger$}	   &N/A&0.62&0.3814&0.66\\\cline{2-7}
							&	\multirow{4}{*}{0.5}
							&\textbf{D-RPCA(C)}					    &0.035&\textbf{0.983}&\textbf{0.017}&\textbf{0.995}\\\cline{3-7}
							&&\textbf{OP$^\dagger$} &0.200&0.940&0.264&0.887\\\cline{3-7}
							&&\textbf{MF}						&N/A&0.980&0.160&0.993\\\cline{3-7}
							&&\textbf{MF$_*^\dagger$}	   &N/A&0.555&0.447&0.442\\\hline
						\end{tabular}
						\label{dl_30:pu_cw}
					}
				\end{subtable} 
				\begin{subtable}{.5\linewidth}
					\captionsetup{font=footnotesize}
					\centering
					\caption{Dictionary by sampling voxels, $d=60$ }
					\scalebox{1}{
						\begin{tabular}{|P{0.5cm}|c|c|c|c|c|}
							\hline 
							\multirow{2}{*}{\textbf{$d$}}  & \multirow{2}{*}{\textbf{Method}}& \multirow{2}{*}{\textbf{Threshold}}&\multicolumn{2}{G|}{\textbf{Performance at best operating point} }& \multirow{2}{*}{\textbf{AUC}}\\ \cline{4-5}
							&&&TPR&FPR&\\\hline
							\multirow{4}{*}{60}
							&\textbf{D-RPCA(C)}					&0.020&\textbf{0.993}&0.022&\textbf{0.994}\\\cline{2-6}
							&\textbf{OP$^\dagger$} &0.250&0.963&0.264&0.907\\\cline{2-6}
								&\textbf{MF}						&N/A&0.980&\textbf{0.011}&0.994\\\cline{2-6}
								&\textbf{MF$^\dagger$}	   &N/A&0.644&0.355&0.700\\\hline
						\end{tabular}
						\label{orig_dict:pu_cw}
					}
					\vspace{0.17cm}
					\caption{Average performance}
					\vspace{-3px}
					\scalebox{0.96}{	
						\begin{tabular}{|P{1.5cm}|c|c|c|c|c|c|}		
							\hline
							\multirow{2}{*}{\textbf{Method}}& \multicolumn{2}{G|}{\textbf{TPR}}& \multicolumn{2}{G|}{\textbf{FPR}}& \multicolumn{2}{G|}{\textbf{AUC}}\\ \cline{2-7}
							&Mean&St.Dev.&Mean&St.Dev.&Mean&St.Dev.\\ \hline
							\textbf{D-RPCA(C)}&	\textbf{0.990}&	\textbf{0.006}&	\textbf{0.015}&	\textbf{0.003}&	\textbf{0.993}&	\textbf{0.002}\\ \hline
							\textbf{OP$^\dagger$}&0.912&0.105&0.302&0.044&0.850&0.098\\ \hline
							\textbf{MF}&0.97&0.025&0.074&0.063&0.984&0.015\\ \hline
							\textbf{MF$^\dagger$}	&0.580&0.064&0.4208&0.065&0.575&0.124\\ \hline
						\end{tabular}
						\label{overall_perf:pu_cw}
					}
				\end{subtable}
				\captionsetup{font=small}
				\vspace*{-10pt}
			\end{table*}